\pdfoutput=1

\documentclass[11pt]{article}

\usepackage[final]{acl}

\usepackage{times}
\usepackage{latexsym}

\usepackage[T1]{fontenc}

\usepackage[utf8]{inputenc}

\usepackage{microtype}

\usepackage{inconsolata}

\usepackage{graphicx}
\usepackage{amsmath}
\usepackage{amsfonts}
\usepackage{multicol}
\usepackage{multirow}
\usepackage{booktabs}
\usepackage{colortbl}
\usepackage{subcaption}
\usepackage{xcolor}
\usepackage{tabularx}
\usepackage{tcolorbox}

\tcbuselibrary{skins}
\tcbuselibrary{breakable}
\usepackage{placeins}
\usepackage{stfloats}
\usepackage{fontawesome5}

\newcommand{\methodname}{\textsc{HyperSteer}}
\newcommand{\gemma}{\texttt{Gemma-2-2B}}
\newcommand{\gpt}{\texttt{gpt-4o-mini}}

\title{\methodname: Activation Steering at Scale with Hypernetworks}

\author{
  Jiuding Sun\thanks{Equal contribution.} \\
  Stanford University \\
  \texttt{sunjd24@stanford.edu}
  \And
  Sidharth Baskaran\footnotemark[1] \\
  Pr(Ai)$^2$R Group \\
  Georgia Institute of Technology \\
  \texttt{sidnbaskaran@gmail.com}
  \And
  Zhengxuan Wu \\
  Stanford University \\
  \texttt{zhengxuan@stanford.edu}
  \AND
  Michael Sklar \\
  Confirm Labs \\
  \texttt{michaelbsklar@gmail.com}
  \And
  Christopher Potts \\
  Stanford University \\
  \texttt{cgpotts@stanford.edu}
  \And
  Atticus Geiger \\
  Pr(Ai)$^2$R Group \\
  \texttt{atticusg@gmail.com}
}

\begin{document}
\maketitle
\begin{abstract}
Steering language models (LMs) by modifying internal activations is a popular approach for controlling text generation. 
Unsupervised dictionary learning methods, e.g., sparse autoencoders, can be scaled to produce many steering vectors, but lack guarantees on the individual efficacy of each vector and control over the coverage of relevant steering tasks.
In contrast, supervised methods for constructing steering vectors are targeted and effective, but require more data collection and training for each additional steering vector produced.
In this work, we introduce \methodname, a family of hypernetwork-based architectures which are trained end-to-end to generate steering vectors conditioned on the natural language steering prompts and the internals of the steered LM. 
In our evaluations, we show that scaling \methodname\ with thousands of steering prompts exceeds the performance of state-of-the-art activation steering methods, even on steering prompts never seen during training. Moreover, \methodname\ performs on par with steering-via-prompting. We release ours at \href{https://github.com/stanfordnlp/axbench}{\faGithub\ \texttt{stanfordnlp/axbench}}.
\end{abstract}

\section{Introduction}

How can the outputs of a language model (LM) be reliably controlled? With instruction-tuned LMs, the standard approach is prompt engineering. 
However, prompting-based approaches face challenges from user jailbreaks, forgotten instructions, and robustness to model misalignment. 
A more aggressive approach is (parameter-efficient) fine-tuning, but this requires modifying or injecting new parameters into the model. 
Activation steering \cite{Giulianelli:2018} occupies a middle ground: it is lightweight to implement (no parameters are modified or added) and can affect the model in ways standard prompting cannot.

Methods for obtaining steering vectors can generally be classified into two groups: \textit{unsupervised}, e.g., Sparse Autoencoders (SAEs) \cite{Hernandez2022, cunningham2023sparseautoencodershighlyinterpretable,bricken2023monosemanticity,gao2024scalingevaluatingsparseautoencoders, marks2025sparse}, which produce a large number of unlabeled steering vectors, and \textit{supervised}, where task-specific steering vectors are either directly trained \cite{wu2024reft, wu2025axbenchsteeringllmssimple} or derived using labeled data \cite{li2023inferencetime,Marks:2018,DBLP:journals/corr/abs-2308-10248, rimsky-etal-2024-steering}. Unsupervised methods are scalable, but lack the ability to create steering vectors for specific tasks.
Alternatively, a supervised approach is effective and targeted, but requires per-task data collection and/or training.

\begin{figure}[t!]
  \centering
  \includegraphics[width=0.5\textwidth]{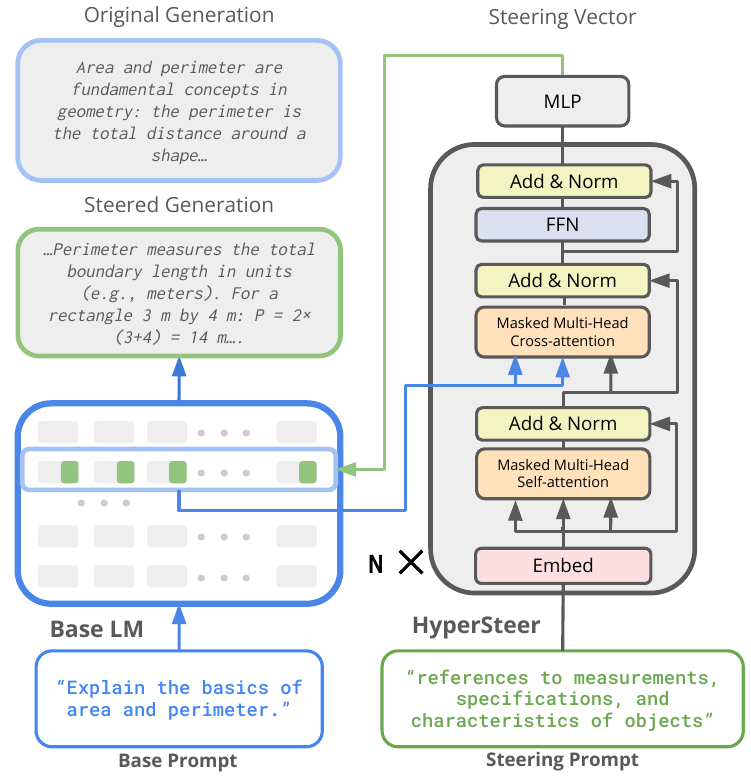}
  \caption{The state-of-the-art \methodname\ Model: A transformer hypernetwork uses self attention to process a steering prompt and uses a cross attention module to read from the residual stream of a base LM run on a second prompt. The hypernetwork outputs a steering vector that is added to the base LM residual stream.}
  \label{fig:overview-slide}
\end{figure}

Our main contribution is \methodname, a suite of end-to-end trainable hypernetwork architectures which learn to generate steering vectors for an instruction-tuned base LM, conditioned on the steering prompt and (optionally) the base LM prompt and internal activations. We train and evaluate on the steering prompts from \textsc{AxBench} \cite{wu2025axbenchsteeringllmssimple}. 
Our best \methodname\ variant outperforms the previous state-of-the-art activation steering method from \textsc{AxBench}, even on held-out steering prompts, never seen during training.
We show steering performance scales logarithmically with number of steering prompts during training.
Surprisingly, we are even able to match the performance of the steering-via-prompting, which outperformed all activation steering methods on AxBench.
In sum, \methodname\ combines the scalability of unsupervised dictionary learning with the targeted control of supervised activation steering.



\begin{figure}
  \centering
  \includegraphics[width=0.475\textwidth]{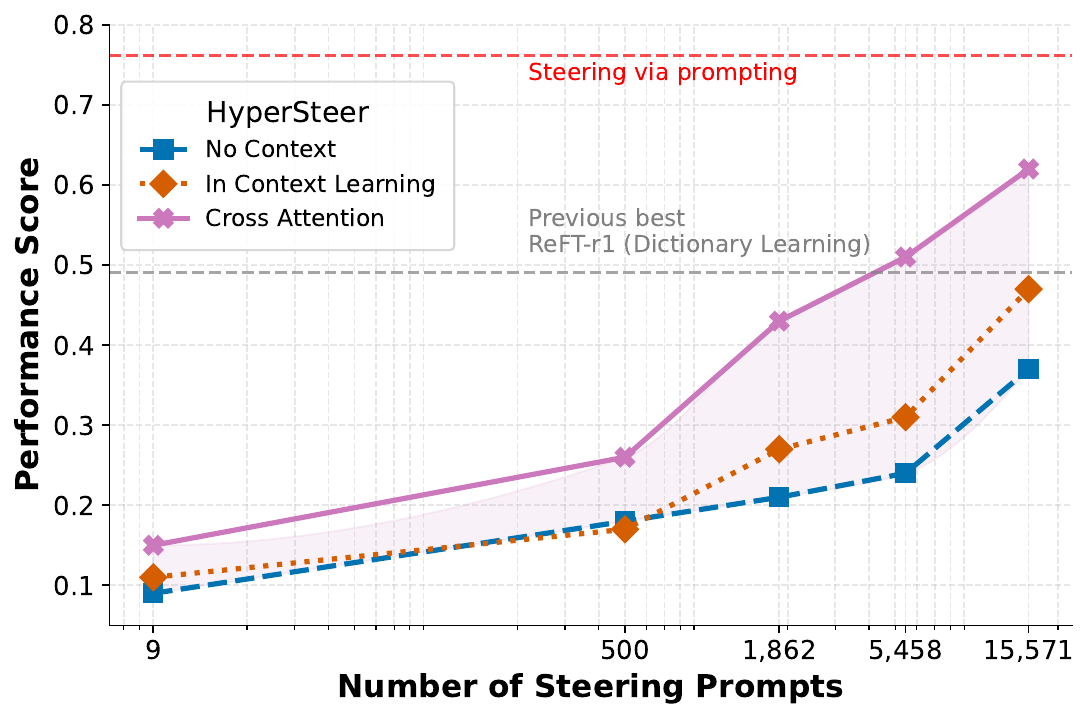}
  \caption{Performance on steering prompts that have \textbf{never been seen during training} for \methodname\ variants as the number of steering prompts used in training increases ($x$-axis in log scale). An exponential increase in training data results in an approximately linear increase in performance. When trained on all $\approx$16k steering prompts in AxBench, the performance of \methodname\ on steering prompts never seen during training surpasses \texttt{ReFT-r1} steering vectors which are trained and evaluated on the same steering prompt.}
  \label{fig:main-scaling}
\end{figure}

\section{Preliminaries}

\paragraph{Activation Steering}
The goal of activation steering is to elicit a certain behavior from a base LM $\mathcal{B}$ run on an base prompt by adding a steering vector to a hidden vector $\mathbf{h}$. Our hypernetwork-based approach to activation steering has the more general goal of taking any base prompt ${x}$, e.g., \textit{Explain how to sort lists of numbers.}, and any steering prompt ${s}$, e.g., \textit{Output using C/C++ programming syntax}, and producing a steering vector $\Delta^{x}_{s}$ added to $\mathbf{h}$.
Denote the steered output:
\begin{equation}
    \hat{\mathbf{y}}_{\textsf{steer}} = \mathcal{B}\bigl({x} ~|~ \mathbf{h} \leftarrow \mathbf{h} + \Delta^{x}_{s}\bigr).
    \label{eq:steering-intervention}
\end{equation}


\paragraph{Hypernetworks}
Prior work has shown hypernetworks to be effective at zero-shot adaptation of language models on a variety of tasks, matching or exceeding comparable methods such as fine-tuning with parameter efficient methods \cite{pmlr-v202-phang23a}.
Formally, a hypernetwork \cite{ha2016hypernetworks} 
is a function $\mathcal{H}: \mathcal{T} \rightarrow \Theta$ that maps tasks $t \in \mathcal{T}$ to parameters $\theta_t = \mathcal{H}(t)$.

For our purposes, the tasks are pairs on input and steering prompts $({x}, {s}) \in \mathcal{T}$ and the output parameters are steering vectors:
\begin{equation}
    \mathcal{H}({x}, {s})= \Delta^x_s \in \mathbb{R}^d.
\end{equation}




\begin{table}[t]
  \small
  \centering
  \resizebox{0.95\columnwidth}{!}{%
    \begin{tabular}{l c c c c}
      \toprule
      \multirow{2}{*}{\textbf{Method}} & \multicolumn{2}{c}{\textbf{Gemma-2-2b}} & \multicolumn{2}{c}{\textbf{Gemma-2-9b}} \\
      & \textbf{Held-out} & \textbf{Held-in} & \textbf{Held-out} & \textbf{Held-in}\\
      \midrule
      \rowcolor{gray!10} \textbf{Prompting} & \textbf{0.762} & 0.731 & \textbf{1.091} & 1.075 \\
      \midrule
      \rowcolor{gray!10} \textbf{Fine-tuning} & & & & \\ 
       LoReFT              & --    & 0.722 & -- & 0.777 \\
       SFT                 & --    & 0.714 & -- & -- \\
       LoRA                & --    & 0.641 & -- & 0.602 \\
      \midrule
      \rowcolor{gray!10} \textbf{Activation Steering} & & & & \\
        ReFT-r1             & --    & 0.509 & -- & 0.630 \\
        DiffMean            & --    & 0.178 & -- & 0.322 \\
        SAE                 & --    & 0.151 & -- & 0.191 \\
        SAE-A               & --    & 0.132 & -- & 0.186 \\
    {\methodname{}} &  & & & \\ 
     – No Context          & 0.373 & 0.512 & 0.633 & 0.751 \\
     – In Context Learning & 0.480 & 0.547 & 0.760 & 0.842 \\
      \rowcolor{blue!15} {– Cross Attention}     & {0.608} & \textbf{0.742} & 0.934 & \textbf{1.091} \\
      \bottomrule
    \end{tabular}%
  }
  \caption{Steering results for baseline methods from AxBench and our three \methodname{} variants, evaluated on \texttt{Concept500-HO} (steering prompts never seen during training) and \texttt{Concept500-HI} (steering prompts seen during training, base prompts unseen during training). The cross attention variant outperforms all other variants by a large margin on held-out evaluation and approaches prompting performance on held-in evaluation, while all variants outperform the \texttt{ReFT-r1} baseline on held-in evaluation. The intervention happened at Layer 20 of both models.}
  \label{tab:consolidated}
\end{table}

\paragraph{Dataset and Evaluation} For all experiments, we use \textsc{AxBench} \cite{wu2025axbenchsteeringllmssimple} as the training dataset and evaluation harness. The training data sets consist of a total of 16,000 steering prompts sourced from \texttt{GemmaScope} Sparse Autoencoder (SAE) feature labels \cite{DBLP:journals/corr/abs-2408-05147} and base prompts sourced from a diverse instruction pool (see App.~\ref{appendix:data} for details). We keep fixed ratios of base prompt to steering prompt ($72:1$ for train; $10:1$ for evaluation).

For evaluation, the steering prompts are applied on a set of different base prompts sourced from AlpacaEval \cite{alpaca_eval}, and a \gpt\ judge model \cite{openai2024gpt4ocard} computes discrete scores in $\{0,1,2\}$ along three dimensions of success: following the base prompt, following the steering prompt, and text fluency. The harmonic mean of these three scores is the final metric, which we refer to as the \textit{steering performance}. 

We evaluate on two datasets: \texttt{Concept500-HI} (held-in), the standard \textsc{AxBench} setting with 500 steering prompts seen during training plus unseen base prompts, and \texttt{Concept500-HO} (held-out), a test set with 500 steering prompts not seen during training. Notably, the only the prompt engineering baseline from AxBench can be evaluated on the held out test set alongside \methodname.


\paragraph{Baselines} We report the fine-tuning, activation steering, and prompting baselines from \textsc{AxBench}.
The state-of-the-art activation steering method \texttt{ReFT-r1} is an important point of comparison for our method. 
See Appendix~\ref{appendix:baseline} for details.


\section{\methodname}
We consider multiple \methodname\ variants, all of which employ a transformer hypernetwork $\mathcal{H}$ with $L$ layers and residual stream representations $\mathbf{h}^t_l$ for each layer $l$ and token $t$ from the steering prompt $\mathbf{s}$ with $T$ tokens. All variants have an MLP module that maps from the residual stream of the last layer and token to a steering vector: 
\begin{equation}
\mathcal{H}(s, x) = \mathsf{MLP}(h^T_L) =\Delta_s^x.
\label{eq:steeringhead}
\end{equation}
We train $\mathcal{B}$ on a language modeling loss using the output $\hat{\mathbf{y}}_{\mathsf{steer}}$ (see Equation~\ref{eq:steering-intervention}) of the base model $\mathcal{B}$ under a steering intervention and an expected output $\mathbf{y}_{\mathsf{label}}$ from AxBench:
\begin{equation}
\mathcal{L}_{\textsf{LM}}(x,s) =  \mathsf{CrossEntropy}(\hat{\mathbf{y}}_{\textsf{steer}},\mathbf{y}_{\textsf{label}}).
\label{eq:sft-loss}
\end{equation}
We consider a range of architectures with incrementally more access to the base LM $\mathcal{L}$ and prompt $x$. 

\paragraph{No context} No access to the base prompt $x$ or the base LM $\mathcal{B}$, meaning $\mathcal H({x},{s})=\mathcal{H}({s})=\Delta_s$.

\paragraph{In Context Learning} This variant appends the base prompt $x$ to the source prompt $s$ and feeds the resulting text into hypernetwork $\mathcal{H}$.

\paragraph{Cross Attention} Our best-performing variant conditions on both the steering prompt $s$ and the internal activations of the base LM $\mathcal{B}$ run on prompt $x$. A cross-attention modules at each layer of the hypernetwork $\mathcal H$ uses the hypernetwork residual stream for attention head queries and outputs and the base LM residual stream at the steering layer $l$ as attention head keys and values (App.~\ref{appendix:crossattn-model}). 
This is a simplification of HyperDAS \cite{sun2025hyperdas}.






\section{Experiments}

We implement \methodname\ building on the \textsc{AxBench} \cite{wu2025axbenchsteeringllmssimple} codebase using 
\texttt{pyvene} \cite{wu-etal-2024-pyvene} to implement steering methods. Training runs are done on a single NVIDIA \texttt{A100-80GB} GPU. Hyperparameter details in \ref{appendix:hparams}. Ourexperiments steer two base LMs, namely, the \texttt{2B} and \texttt{9B} variants of \texttt{Gemma-2}\; instruction-tuned. \cite{DBLP:journals/corr/abs-2408-00118}, with its parameters frozen during training. The hypernetworks used are unfrozen copies of the base model with later layers removed. We train each \methodname\ variant on steering prompt datasets ranging from 10 up to the full $\approx16 000$ steering prompts available in \textsc{AxBench}. 

\begin{figure}
    \centering
    \includegraphics[width=0.9\linewidth]{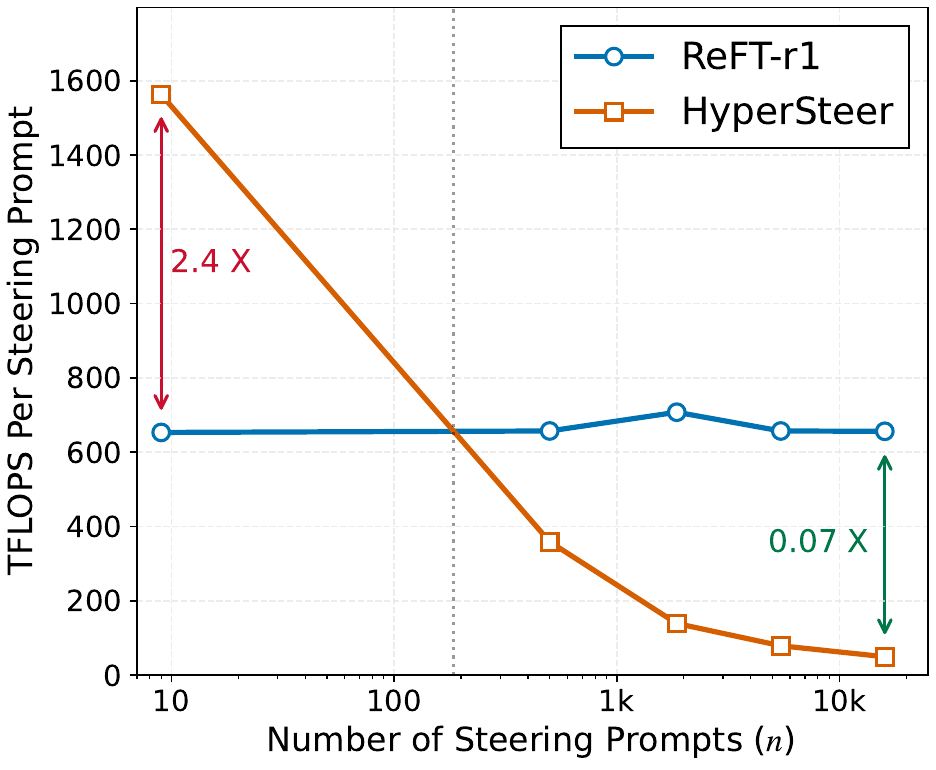}
    \caption{
        As the number of steering prompts in our training dataset increases, the teraFLOPs (TFLOPS) required to attain a similar loss on a \textit{held-in} evaluation set (steering prompts seen during training, but base prompt unseen during training) decreases for our best \methodname\ variant (cross attention). This value is approximately constant for our dictionary learning baseline of \texttt{ReFT-r1}. See Appendix~\ref{appendix:flops} for details.
    }
    \label{fig:flop-limits}
\end{figure}

\paragraph{Generalization Results}
We evaluate \methodname\ on \textsc{Axbench}'s \texttt{Concept500-HI} and \texttt{Concept500-HO} and report results in Table \ref{tab:consolidated}. \textbf{our cross-attention \methodname\ variant performs better on unseen steering prompts than every supervised activation steering baseline trained and evaluated on the same steering prompt.} However, \methodname\ falls slightly behind prompting and the best fine-tuning baselines.
Figure \ref{fig:main-scaling} shows that the cross-attention variant outperforms the other architectures at every dataset scale.

\paragraph{Compute efficiency}
We study the efficacy of \methodname\ with respect to the FLOPs needed to maintain the evaluation loss on a held-in dataset as we increase the training data used. We compare against the state-of-the-art supervised activation steering method \texttt{ReFT-r1}. Observe in Figure~\ref{fig:flop-limits}), that as training data increases \methodname\ becomes much more economical than supervised activation steering. See details in Appendix~\ref{appendix:flops}.

\paragraph{Ablation Study}

\begin{table}
    \small
    \centering
    \resizebox{\columnwidth}{!}{\begin{tabular}{l c c}
        \toprule
        \textbf{\methodname\; (Cross Attention)} & Held-in & Held-out \\
        \midrule
        \textit{Initialization} & \\
         Random & 0.601 & 0.582 \\
         \rowcolor{blue!15} Pretrained & 0.733 & \textbf{0.623} \\
         \midrule
         \textit{Decoder Blocks} & \\
         $N=2$ & 0.707 & 0.549 \\
         $N=4$ & 0.713 & 0.597 \\
         $N=8$ & 0.721 & 0.610 \\
         \rowcolor{blue!15} $N=20$ & 0.733 & \textbf{0.623} \\
        \midrule
        \textbf{\methodname\; (No Context)} &  & \\
        \midrule
        \textit{Training Objective} & \\
        Reconstruction Loss & 0.511 & 0.375 \\ 
        \rowcolor{blue!15}Language Modeling Loss & 0.517 & \textbf{0.340} \\
         \bottomrule
    \end{tabular}}
    \caption{Ablation study using \gemma\ on architecture choices of \methodname\ after 1 epoch of training and evaluation on a small test set \texttt{Concept10}. We find that pre-trained initialization of the cross-attention architecture improves performance in both held-in and held-out scenarios, and in both cases performance improves with the number of hypernetwork decoder blocks. For the no-context hypernetworks which do not condition steering vectors on input prompts, reconstructing ground truth vectors is comparable to end-to-end training with a language modeling objective.}
    \label{tab:ablations}
\end{table}

We show results for various ablation studies on the cross-attention variant of \methodname\ in Table~\ref{tab:ablations}.
We randomly initialize the \texttt{Gemma2-2B} hypernetwork and find that pretrained parameters provide a significant performance boost (+0.112 steering score). 
    We also remove a number of hypernetwork decoder blocks in the range $N=\{2,4,8,20\}$ and adjust learning rate accordingly: $\mathrm{lr}(n) = 8\times10^{-5} \cdot \sqrt{\frac{20}{n}}$. Increased depth results in incremental improvements to steering performance on held in and held out test sets. However, notably the number of decoder blocks has a greater impact on generalization to steering prompts unseen in training (+0.07) compared to steering prompts unseen in training  (+0.03).

We also perform an ablation on the no context \methodname\ variant where we train the hypernetwork to reconstruct the steering vectors constructed by the original AxBench ReFT baselines. Given a steering vector $\mathcal{H}(s,x) = \mathcal{H}(s) = \Delta_s$ and a gold-label steering vector $\Delta^*_s$ the loss is 
$$\mathcal{L}_{\textsf{recon}}(s) = 1-\mathsf{CosSim}(\Delta_s, \Delta^*_s) + ||\Delta_s - \Delta^*_s||_2^2.$$
The two loss terms are roughly comparable, so we use language modeling.

\section{Qualitative Analyses}
We generate 2500 steering vectors using base and steering prompts from our held-out test data.

\paragraph{Geometric visualization of steering vectors}
We analyze steering vectors generated by \methodname\; (Cross Attention) using t-SNE \cite{JMLR:v9:vandermaaten08a} and PCA (2 components) to find geometric structure among steering vectors (see Fig. \ref{fig:tsne-concept10} and \ref{fig:pca-concept10} in App. \ref{appendix:geom-structure}).

\paragraph{Pairwise similarity of steering vectors}
We compute pairwise cosine similarities of steering vectors on both in-context (reconstruction) and cross attention models to understand how conditioning on the input prompt affects semantics.
The cross-attention variant (Figure~\ref{fig:sim-ca-concept10} in App. \ref{appendix:geom-structure}) yields high within-concept alignment but still shows off-diagonal similarities driven by shared prompt templates and linguistic structure. In contrast, the no-context variant (Figure~\ref{fig:sim-nc-concept10} in \ref{appendix:geom-structure}), conditioning on steering prompt only, produces much weaker off-diagonal alignment. We find that cross-attention’s residual inter-concept similarity is weakened by this additional conditioning, but not at the cost of steering performance. Initial experiments to determine if geometric structure emerges among steering vectors sharing a concept yielded a negative. This is likely due to high semantic similarity of the prompts used in our evaluation pipeline.

\section{Conclusion}

Both held-in and held-out evaluations indicate that \methodname\ is a scalable and effective approach for steering language models. In particular, \methodname\ (Cross Attention), our best-performing variant, achieves significantly stronger performance on held-out prompts—improving further with dataset scale. It also outperforms all activation steering baselines on held-in evaluations. Without modifying model parameters, our method narrows the performance gap with fine-tuning and prompting. Finally, we demonstrate that \methodname\ becomes increasingly compute-efficient as data scale increases, achieving the same held-out loss with fewer training updates.


\section{Limitations}
\paragraph{Data} A key limitation of our approach is the limited scope and quantity of the concept datasets. Using data with concepts of much greater complexity and difficulty from a model steering perspective would likely improve model performance and help make evaluation more robust. We also note that quality and robustness of concepts is bounded by the \texttt{GemmaScope} feature labels used to derive them, and collecting data from humans or other high quality sources is a feasible alternative. This is a key research priority we emphasize for future work.


\paragraph{Steering Sites} All experiments in our work are limited to intervening on the \textit{residual stream} activations of the base LM. There are other potentially more performant sites for intervention, including various points of the decoder block and during the attention computation. We also adopt the convention of prior work to intervene at all token positions; exploring more targeted interventions could reduce detrimental off-target steering effects and improve the overall steering score.

\paragraph{Compute} Compared to supervised dictionary learning, the compute requirements of training a hypernetwork are large, as the number of trainable parameters significantly exceeds a \texttt{ReFT-r1}.

\paragraph{Model Scale} Due to to compute constraints we only experimented with \gemma\; architectures, which are worse instruction followers and in-context learners than the leading open source models with many more parameters. Training on models at a variety of scale would help cement \methodname\;'s strong steering performance against the improved in-context learning ability of larger LMs.

\paragraph{Open Source Models} Our approach requires white-box access to a model's internals in order to use steering vectors, a limitation prompting does not encounter. Hence, we rely on the existence of sufficiently capable open source models as a basis for our research.


\section{Ethical Considerations}
We present this work with the intention that \methodname\ is a powerful tool for steering models away from producing harmful responses and better tailor outputs to downstream tasks. However, we acknowledge that model steering can also be used by bad actors as a tool to circumvent a target models's existing safety mechanisms or bias models towards misleading outputs or malicious persuasion. Hence, \methodname\ and hence steering vectors should be used responsibly and audited to prevent such issues from arising, and having a human-in-the-loop system could help mitigate some of these concerns.

\section{Acknowledgments}

\paragraph{AI Usage}
We use closed-source LLMs from OpenAI as a critical part of our work: synthetic concept data generation and evaluation pipelines utilize \gpt\; to generate ground truth labels and judge responses according to criteria respectively.
\paragraph{Other} This research was in part supported by a grant from Open Philanthropy. We thank Aryaman Arora, Róbert Csordás, and Qinan Yu for constant and extremely helpful feedback during the discussion. 



\bibliography{custom}

\begin{thebibliography}{24}
\providecommand{\natexlab}[1]{#1}

\bibitem[{Bricken et~al.(2023)Bricken, Templeton, Batson, Chen, Jermyn, Conerly, Turner, Anil, Denison, Askell, Lasenby, Wu, Kravec, Schiefer, Maxwell, Joseph, Hatfield-Dodds, Tamkin, Nguyen, McLean, Burke, Hume, Carter, Henighan, and Olah}]{bricken2023monosemanticity}
Trenton Bricken, Adly Templeton, Joshua Batson, Brian Chen, Adam Jermyn, Tom Conerly, Nick Turner, Cem Anil, Carson Denison, Amanda Askell, Robert Lasenby, Yifan Wu, Shauna Kravec, Nicholas Schiefer, Tim Maxwell, Nicholas Joseph, Zac Hatfield-Dodds, Alex Tamkin, Karina Nguyen, and 6 others. 2023.
\newblock Towards monosemanticity: Decomposing language models with dictionary learning.
\newblock \emph{Transformer Circuits Thread}.
\newblock Https://transformer-circuits.pub/2023/monosemantic-features/index.html.

\bibitem[{Cobbe et~al.(2021)Cobbe, Kosaraju, Bavarian, Chen, Jun, Kaiser, Plappert, Tworek, Hilton, Nakano, Hesse, and Schulman}]{cobbe2021gsm8k}
Karl Cobbe, Vineet Kosaraju, Mohammad Bavarian, Mark Chen, Heewoo Jun, Lukasz Kaiser, Matthias Plappert, Jerry Tworek, Jacob Hilton, Reiichiro Nakano, Christopher Hesse, and John Schulman. 2021.
\newblock Training verifiers to solve math word problems.
\newblock \emph{arXiv preprint arXiv:2110.14168}.

\bibitem[{Conover et~al.(2023)Conover, Hayes, Mathur, Xie, Wan, Shah, Ghodsi, Wendell, Zaharia, and Xin}]{DatabricksBlog2023DollyV2}
Mike Conover, Matt Hayes, Ankit Mathur, Jianwei Xie, Jun Wan, Sam Shah, Ali Ghodsi, Patrick Wendell, Matei Zaharia, and Reynold Xin. 2023.
\newblock \href {https://www.databricks.com/blog/2023/04/12/dolly-first-open-commercially-viable-instruction-tuned-llm} {Free dolly: Introducing the world's first truly open instruction-tuned llm}.

\bibitem[{Cunningham et~al.(2023)Cunningham, Ewart, Riggs, Huben, and Sharkey}]{cunningham2023sparseautoencodershighlyinterpretable}
Hoagy Cunningham, Aidan Ewart, Logan Riggs, Robert Huben, and Lee Sharkey. 2023.
\newblock \href {https://arxiv.org/abs/2309.08600} {Sparse autoencoders find highly interpretable features in language models}.
\newblock \emph{Preprint}, arXiv:2309.08600.

\bibitem[{Gao et~al.(2024)Gao, la~Tour, Tillman, Goh, Troll, Radford, Sutskever, Leike, and Wu}]{gao2024scalingevaluatingsparseautoencoders}
Leo Gao, Tom~Dupré la~Tour, Henk Tillman, Gabriel Goh, Rajan Troll, Alec Radford, Ilya Sutskever, Jan Leike, and Jeffrey Wu. 2024.
\newblock \href {https://arxiv.org/abs/2406.04093} {Scaling and evaluating sparse autoencoders}.
\newblock \emph{Preprint}, arXiv:2406.04093.

\bibitem[{Giulianelli et~al.(2018)Giulianelli, Harding, Mohnert, Hupkes, and Zuidema}]{Giulianelli:2018}
Mario Giulianelli, Jack Harding, Florian Mohnert, Dieuwke Hupkes, and Willem~H. Zuidema. 2018.
\newblock \href {https://doi.org/10.18653/V1/W18-5426} {Under the hood: Using diagnostic classifiers to investigate and improve how language models track agreement information}.
\newblock In \emph{Proceedings of the Workshop: Analyzing and Interpreting Neural Networks for NLP, BlackboxNLP EMNLP 2018, Brussels, Belgium, November 1, 2018}, pages 240--248. Association for Computational Linguistics.

\bibitem[{Ha et~al.(2016)Ha, Dai, and Le}]{ha2016hypernetworks}
David Ha, Andrew Dai, and Quoc~V. Le. 2016.
\newblock \href {https://arxiv.org/abs/1609.09106} {Hypernetworks}.
\newblock \emph{Preprint}, arXiv:1609.09106.

\bibitem[{Hernandez et~al.(2022)Hernandez, Schwettmann, Bau, Bagashvili, Torralba, and Andreas}]{Hernandez2022}
Evan Hernandez, Sarah Schwettmann, David Bau, Teona Bagashvili, Antonio Torralba, and Jacob Andreas. 2022.
\newblock \href {https://openreview.net/forum?id=NudBMY-tzDr} {Natural language descriptions of deep visual features}.
\newblock In \emph{ICLR}.

\bibitem[{Hu et~al.(2022)Hu, Shen, Wallis, Allen-Zhu, Li, Wang, Wang, Chen et~al.}]{hu2022lora}
Edward~J Hu, Yelong Shen, Phillip Wallis, Zeyuan Allen-Zhu, Yuanzhi Li, Shean Wang, Lu~Wang, Weizhu Chen, and 1 others. 2022.
\newblock Lora: Low-rank adaptation of large language models.
\newblock \emph{ICLR}, 1(2):3.

\bibitem[{Li et~al.(2023{\natexlab{a}})Li, Patel, Vi{\'e}gas, Pfister, and Wattenberg}]{li2023inferencetime}
Kenneth Li, Oam Patel, Fernanda Vi{\'e}gas, Hanspeter Pfister, and Martin Wattenberg. 2023{\natexlab{a}}.
\newblock \href {https://openreview.net/forum?id=aLLuYpn83y} {Inference-time intervention: Eliciting truthful answers from a language model}.
\newblock In \emph{Thirty-seventh Conference on Neural Information Processing Systems}.

\bibitem[{Li et~al.(2023{\natexlab{b}})Li, Zhang, Dubois, Taori, Gulrajani, Guestrin, Liang, and Hashimoto}]{alpaca_eval}
Xuechen Li, Tianyi Zhang, Yann Dubois, Rohan Taori, Ishaan Gulrajani, Carlos Guestrin, Percy Liang, and Tatsunori~B. Hashimoto. 2023{\natexlab{b}}.
\newblock Alpacaeval: An automatic evaluator of instruction-following models.
\newblock \url{https://github.com/tatsu-lab/alpaca_eval}.

\bibitem[{Lieberum et~al.(2024)Lieberum, Rajamanoharan, Conmy, Smith, Sonnerat, Varma, Kram{\'{a}}r, Dragan, Shah, and Nanda}]{DBLP:journals/corr/abs-2408-05147}
Tom Lieberum, Senthooran Rajamanoharan, Arthur Conmy, Lewis Smith, Nicolas Sonnerat, Vikrant Varma, J{\'{a}}nos Kram{\'{a}}r, Anca~D. Dragan, Rohin Shah, and Neel Nanda. 2024.
\newblock \href {https://doi.org/10.48550/ARXIV.2408.05147} {Gemma scope: Open sparse autoencoders everywhere all at once on gemma 2}.
\newblock \emph{CoRR}, abs/2408.05147.

\bibitem[{Marks et~al.(2025)Marks, Rager, Michaud, Belinkov, Bau, and Mueller}]{marks2025sparse}
Samuel Marks, Can Rager, Eric~J Michaud, Yonatan Belinkov, David Bau, and Aaron Mueller. 2025.
\newblock \href {https://openreview.net/forum?id=I4e82CIDxv} {Sparse feature circuits: Discovering and editing interpretable causal graphs in language models}.
\newblock In \emph{The Thirteenth International Conference on Learning Representations}.

\bibitem[{Marks and Tegmark(2023)}]{Marks:2018}
Samuel Marks and Max Tegmark. 2023.
\newblock \href {https://doi.org/10.48550/ARXIV.2310.06824} {The geometry of truth: Emergent linear structure in large language model representations of true/false datasets}.
\newblock \emph{CoRR}, abs/2310.06824.

\bibitem[{OpenAI et~al.(2024)OpenAI, :, Hurst, Lerer, Goucher, Perelman, Ramesh, Clark, Ostrow, Welihinda, Hayes, Radford, Mądry, Baker-Whitcomb, Beutel, Borzunov, Carney, Chow, Kirillov, Nichol, Paino, Renzin, Passos, Kirillov, Christakis, Conneau, Kamali, Jabri, Moyer, Tam, Crookes, Tootoochian, Tootoonchian, Kumar, Vallone, Karpathy, Braunstein, Cann, Codispoti, Galu, Kondrich, Tulloch, Mishchenko, Baek, Jiang, Pelisse, Woodford, Gosalia, Dhar, Pantuliano, Nayak, Oliver, Zoph, Ghorbani, Leimberger, Rossen, Sokolowsky, Wang, Zweig, Hoover, Samic, McGrew, Spero, Giertler, Cheng, Lightcap, Walkin, Quinn, Guarraci, Hsu, Kellogg, Eastman, Lugaresi, Wainwright, Bassin, Hudson, Chu, Nelson, Li, Shern, Conger, Barette, Voss, Ding, Lu, Zhang, Beaumont, Hallacy, Koch, Gibson, Kim, Choi, McLeavey, Hesse, Fischer, Winter, Czarnecki, Jarvis, Wei, Koumouzelis, Sherburn, Kappler, Levin, Levy, Carr, Farhi, Mely, Robinson, Sasaki, Jin, Valladares, Tsipras, Li, Nguyen, Findlay, Oiwoh, Wong, Asdar, Proehl, Yang, Antonow,
  Kramer, Peterson, Sigler, Wallace, Brevdo, Mays, Khorasani, Such, Raso, Zhang, von Lohmann, Sulit, Goh, Oden, Salmon, Starace, Brockman, Salman, Bao, Hu, Wong, Wang, Schmidt, Whitney, Jun, Kirchner, de~Oliveira~Pinto, Ren, Chang, Chung, Kivlichan, O'Connell, O'Connell, Osband, Silber, Sohl, Okuyucu, Lan, Kostrikov, Sutskever, Kanitscheider, Gulrajani, Coxon, Menick, Pachocki, Aung, Betker, Crooks, Lennon, Kiros, Leike, Park, Kwon, Phang, Teplitz, Wei, Wolfe, Chen, Harris, Varavva, Lee, Shieh, Lin, Yu, Weng, Tang, Yu, Jang, Candela, Beutler, Landers, Parish, Heidecke, Schulman, Lachman, McKay, Uesato, Ward, Kim, Huizinga, Sitkin, Kraaijeveld, Gross, Kaplan, Snyder, Achiam, Jiao, Lee, Zhuang, Harriman, Fricke, Hayashi, Singhal, Shi, Karthik, Wood, Rimbach, Hsu, Nguyen, Gu-Lemberg, Button, Liu, Howe, Muthukumar, Luther, Ahmad, Kai, Itow, Workman, Pathak, Chen, Jing, Guy, Fedus, Zhou, Mamitsuka, Weng, McCallum, Held, Ouyang, Feuvrier, Zhang, Kondraciuk, Kaiser, Hewitt, Metz, Doshi, Aflak, Simens, Boyd,
  Thompson, Dukhan, Chen, Gray, Hudnall, Zhang, Aljubeh, Litwin, Zeng, Johnson, Shetty, Gupta, Shah, Yatbaz, Yang, Zhong, Glaese, Chen, Janner, Lampe, Petrov, Wu, Wang, Fradin, Pokrass, Castro, de~Castro, Pavlov, Brundage, Wang, Khan, Murati, Bavarian, Lin, Yesildal, Soto, Gimelshein, Cone, Staudacher, Summers, LaFontaine, Chowdhury, Ryder, Stathas, Turley, Tezak, Felix, Kudige, Keskar, Deutsch, Bundick, Puckett, Nachum, Okelola, Boiko, Murk, Jaffe, Watkins, Godement, Campbell-Moore, Chao, McMillan, Belov, Su, Bak, Bakkum, Deng, Dolan, Hoeschele, Welinder, Tillet, Pronin, Tillet, Dhariwal, Yuan, Dias, Lim, Arora, Troll, Lin, Lopes, Puri, Miyara, Leike, Gaubert, Zamani, Wang, Donnelly, Honsby, Smith, Sahai, Ramchandani, Huet, Carmichael, Zellers, Chen, Chen, Nigmatullin, Cheu, Jain, Altman, Schoenholz, Toizer, Miserendino, Agarwal, Culver, Ethersmith, Gray, Grove, Metzger, Hermani, Jain, Zhao, Wu, Jomoto, Wu, Shuaiqi, Xia, Phene, Papay, Narayanan, Coffey, Lee, Hall, Balaji, Broda, Stramer, Xu, Gogineni,
  Christianson, Sanders, Patwardhan, Cunninghman, Degry, Dimson, Raoux, Shadwell, Zheng, Underwood, Markov, Sherbakov, Rubin, Stasi, Kaftan, Heywood, Peterson, Walters, Eloundou, Qi, Moeller, Monaco, Kuo, Fomenko, Chang, Zheng, Zhou, Manassra, Sheu, Zaremba, Patil, Qian, Kim, Cheng, Zhang, He, Zhang, Jin, Dai, and Malkov}]{openai2024gpt4ocard}
OpenAI, :, Aaron Hurst, Adam Lerer, Adam~P. Goucher, Adam Perelman, Aditya Ramesh, Aidan Clark, AJ~Ostrow, Akila Welihinda, Alan Hayes, Alec Radford, Aleksander Mądry, Alex Baker-Whitcomb, Alex Beutel, Alex Borzunov, Alex Carney, Alex Chow, Alex Kirillov, and 401 others. 2024.
\newblock \href {https://arxiv.org/abs/2410.21276} {Gpt-4o system card}.
\newblock \emph{Preprint}, arXiv:2410.21276.

\bibitem[{Phang et~al.(2023)Phang, Mao, He, and Chen}]{pmlr-v202-phang23a}
Jason Phang, Yi~Mao, Pengcheng He, and Weizhu Chen. 2023.
\newblock \href {https://proceedings.mlr.press/v202/phang23a.html} {{H}yper{T}uning: Toward adapting large language models without back-propagation}.
\newblock In \emph{Proceedings of the 40th International Conference on Machine Learning}, volume 202 of \emph{Proceedings of Machine Learning Research}, pages 27854--27875. PMLR.

\bibitem[{Rimsky et~al.(2024)Rimsky, Gabrieli, Schulz, Tong, Hubinger, and Turner}]{rimsky-etal-2024-steering}
Nina Rimsky, Nick Gabrieli, Julian Schulz, Meg Tong, Evan Hubinger, and Alexander Turner. 2024.
\newblock \href {https://doi.org/10.18653/v1/2024.acl-long.828} {Steering llama 2 via contrastive activation addition}.
\newblock In \emph{Proceedings of the 62nd Annual Meeting of the Association for Computational Linguistics (Volume 1: Long Papers)}, pages 15504--15522, Bangkok, Thailand. Association for Computational Linguistics.

\bibitem[{Rivi{\`{e}}re et~al.(2024)Rivi{\`{e}}re, Pathak, Sessa, Hardin, Bhupatiraju, Hussenot, Mesnard, Shahriari, Ram{\'{e}}, Ferret, Liu, Tafti, Friesen, Casbon, Ramos, Kumar, Lan, Jerome, Tsitsulin, Vieillard, Stanczyk, Girgin, Momchev, Hoffman, Thakoor, Grill, Neyshabur, Bachem, Walton, Severyn, Parrish, Ahmad, Hutchison, Abdagic, Carl, Shen, Brock, Coenen, Laforge, Paterson, Bastian, Piot, Wu, Royal, Chen, Kumar, Perry, Welty, Choquette{-}Choo, Sinopalnikov, Weinberger, Vijaykumar, Rogozinska, Herbison, Bandy, Wang, Noland, Moreira, Senter, Eltyshev, Visin, Rasskin, Wei, Cameron, Martins, Hashemi, Klimczak{-}Plucinska, Batra, Dhand, Nardini, Mein, Zhou, Svensson, Stanway, Chan, Zhou, Carrasqueira, Iljazi, Becker, Fernandez, van Amersfoort, Gordon, Lipschultz, Newlan, Ji, Mohamed, Badola, Black, Millican, McDonell, Nguyen, Sodhia, Greene, Sj{\"{o}}sund, Usui, Sifre, Heuermann, Lago, and McNealus}]{DBLP:journals/corr/abs-2408-00118}
Morgane Rivi{\`{e}}re, Shreya Pathak, Pier~Giuseppe Sessa, Cassidy Hardin, Surya Bhupatiraju, L{\'{e}}onard Hussenot, Thomas Mesnard, Bobak Shahriari, Alexandre Ram{\'{e}}, Johan Ferret, Peter Liu, Pouya Tafti, Abe Friesen, Michelle Casbon, Sabela Ramos, Ravin Kumar, Charline~Le Lan, Sammy Jerome, Anton Tsitsulin, and 80 others. 2024.
\newblock \href {https://doi.org/10.48550/ARXIV.2408.00118} {Gemma 2: Improving open language models at a practical size}.
\newblock \emph{CoRR}, abs/2408.00118.

\bibitem[{Sun et~al.(2025)Sun, Huang, Baskaran, D'Oosterlinck, Potts, Sklar, and Geiger}]{sun2025hyperdas}
Jiuding Sun, Jing Huang, Sidharth Baskaran, Karel D'Oosterlinck, Christopher Potts, Michael Sklar, and Atticus Geiger. 2025.
\newblock \href {https://openreview.net/forum?id=6fDjUoEQvm} {Hyper{DAS}: Towards automating mechanistic interpretability with hypernetworks}.
\newblock In \emph{The Thirteenth International Conference on Learning Representations}.

\bibitem[{Turner et~al.(2023)Turner, Thiergart, Udell, Leech, Mini, and MacDiarmid}]{DBLP:journals/corr/abs-2308-10248}
Alexander~Matt Turner, Lisa Thiergart, David Udell, Gavin Leech, Ulisse Mini, and Monte MacDiarmid. 2023.
\newblock \href {https://doi.org/10.48550/ARXIV.2308.10248} {Activation addition: Steering language models without optimization}.
\newblock \emph{CoRR}, abs/2308.10248.

\bibitem[{van~der Maaten and Hinton(2008)}]{JMLR:v9:vandermaaten08a}
Laurens van~der Maaten and Geoffrey Hinton. 2008.
\newblock \href {http://jmlr.org/papers/v9/vandermaaten08a.html} {Visualizing data using t-sne}.
\newblock \emph{Journal of Machine Learning Research}, 9(86):2579--2605.

\bibitem[{Wu et~al.(2025)Wu, Arora, Geiger, Wang, Huang, Jurafsky, Manning, and Potts}]{wu2025axbenchsteeringllmssimple}
Zhengxuan Wu, Aryaman Arora, Atticus Geiger, Zheng Wang, Jing Huang, Dan Jurafsky, Christopher~D. Manning, and Christopher Potts. 2025.
\newblock \href {https://arxiv.org/abs/2501.17148} {{AxBench}: Steering llms? even simple baselines outperform sparse autoencoders}.
\newblock \emph{Preprint}, arXiv:2501.17148.

\bibitem[{Wu et~al.(2024{\natexlab{a}})Wu, Arora, Wang, Geiger, Jurafsky, Manning, and Potts}]{wu2024reft}
Zhengxuan Wu, Aryaman Arora, Zheng Wang, Atticus Geiger, Dan Jurafsky, Christopher~D Manning, and Christopher Potts. 2024{\natexlab{a}}.
\newblock \href {https://openreview.net/forum?id=fykjplMc0V} {Re{FT}: Representation finetuning for language models}.
\newblock In \emph{The Thirty-eighth Annual Conference on Neural Information Processing Systems}.

\bibitem[{Wu et~al.(2024{\natexlab{b}})Wu, Geiger, Arora, Huang, Wang, Goodman, Manning, and Potts}]{wu-etal-2024-pyvene}
Zhengxuan Wu, Atticus Geiger, Aryaman Arora, Jing Huang, Zheng Wang, Noah Goodman, Christopher Manning, and Christopher Potts. 2024{\natexlab{b}}.
\newblock \href {https://aclanthology.org/2024.naacl-demo.16} {pyvene: A library for understanding and improving {P}y{T}orch models via interventions}.
\newblock In \emph{Proceedings of the 2024 Conference of the North American Chapter of the Association for Computational Linguistics: Human Language Technologies (Volume 3: System Demonstrations)}, pages 158--165, Mexico City, Mexico. Association for Computational Linguistics.

\end{thebibliography}

\appendix

\section{Appendix}
\label{sec:appendix}
\subsection{Future Directions}
\label{app:futuredirections}

\paragraph{Large-scale concept data} Prior works \cite{pmlr-v202-phang23a} explore pre-training the architecture prior to using task-specific data. Since \methodname\; uses a pre-trained model as a starting point, we postulate that sourcing significantly more data with more concepts of varying and complexity would allow us to test the architecture's limits.
\paragraph{Generating other parameter types} Due to compute constraints and our focus on activation steering, we did not explore generating other types of parameter-efficient modulations, including rank-$r$ generalizations such as ReFT \cite{wu2024reft} or LoRA \cite{hu2022lora} adapters. Such generalizations could potentially be more expressive and allow the hypernetwork to adapt language models to more difficult tasks.
\paragraph{Architecture optimizations} \methodname\; (Cross Attention) is a parameter-dense transformer model itself. More efficient alternatives could bridge the gap with the dictionary learning baseline and scale up the approach given a limited compute budget.

\subsection{Hyperparameter Details}

For the \texttt{ReFT-r1} baseline, we use the default hyperparameters and settings from \textsc{AxBench}. We reduce the number of layer for the cross attention variant to match the total number of parameter with the cross-attention model.

\begin{table}[ht!]
  \small
  \centering
  \begin{tabularx}{\columnwidth}{l c c}
    \toprule
    \textbf{Hyperparameters} 
      & \textbf{\textit{Cross Attention}} &  \textbf{\textit{Other variants}} \\
    \midrule
    \texttt{Gemma-2-2b} & \\
    Batch size  
      & 12 & 6 \\
    LR  
      & 8e-5 & 8e-5  \\
    N epoch  
      & 3 & 3 \\
    Layer
      & 22 & 26 \\
    Cross-attention heads 
    & 8 & -- \\     
    \midrule
    \texttt{Gemma-2-9b} & \\
    Batch size  
      & 4 & 6 \\
    LR  
      & 5e-6 & 5e-6  \\
    N epoch  
      & 3 & 3 \\
    Layer
      & 34 & 42 \\
    Cross-attention heads 
    & 8 & -- \\     
    \bottomrule
  \end{tabularx}
  \caption{Hyperparameter settings for each variant. For all the unmentioned details such as hidden dimension we use the default configuration of \texttt{Gemma-2-2b} and \texttt{Gemma-2-9b}.}
  \label{table:hparams}
\end{table}

\label{appendix:hparams}

\subsection{Cross-Attention Architecture Details}

\label{appendix:crossattn-model}

A token sequence $s$ of length $|s|$ representing the concept for steering is encoded as ${\bf h}_0 = \mathrm{Emb}_\Phi({\bf x}) \in \mathbb{R}^{|s|\times d}$. For clarity, we refer to this as the zeroth layer of the residual stream for the hypernetwork $\mathcal H_\Phi$.

This precedes $N$ decoder blocks. Each block contains the standard multi-headed self-attention ($\mathsf{MHA}$) feed-forward layer ($\mathsf{FFN}$), and a multi-headed cross-attention module to include information from $\mathrm{LM}_\mathrm{base}$. 
 Let ${\bf S}\in \mathbb R^{|s|\times d}$ and ${\bf X}^{(p-1)}\in \mathbb R^{|s|\times d}$ be the incoming residual stream.
In the \(p\)‐th block, we compute:
\[
\begin{aligned}
  {\bf X}^{(p)}
    &:= \mathsf{MHA}\bigl(
         \mathbf Q = \bf X^{(p-1)},\;
         \mathbf K = \bf X^{(p-1)},\\[-2pt]
    &\qquad\quad
         \mathbf V = \bf X^{(p-1)}
       \bigr),\\[-2pt]
  {\bf X}^{(p)}
    &:= \mathsf{MHA}\bigl(
         \mathbf Q = \bf S,\;
         \mathbf K = {\bf X}^{(p)},\\[-2pt]
    &\qquad\quad
         \mathbf V = {\bf X}^{(p)}
       \bigr),\\[-2pt]
  {\bf X}^{(p)}
    &:= \mathsf{LayerNorm}\Bigl(
         {\bf X}^{(p-1)} + \mathsf{FFN}\bigl({\bf X}^{(p)}\bigr)
       \Bigr).
\end{aligned}
\]

We initialize the self-attention MHA blocks from the pre-trained \texttt{Gemma-2} base model and the cross attention blocks according to the default PyTorch weight initialization scheme.

\subsubsection{Model Size}

\label{appendix:param-efficiency}

Our largest \methodname\; (cross attention) architecture has 22 modified decoder blocks, and has $\approx 0.998$ times the parameters as \gemma, which has 26 standard decoder blocks. Each cross attention decoder block has $\approx 1.18$ times as many parameters as a standard \gemma\; decoder block.

\subsubsection{Detailed TFLOPs Analysis}
\label{appendix:flops}

We demonstrate that the number of TFLOPs to reach optimal steering performance decays with the number of concepts in the dataset, or steering prompts used. Thus, as we scale up steering data \methodname\; becomes an efficient and superior alternative to the supervised dictionary learning baseline for steering language models. We focus our best method, the cross attention architecture, for this analysis.

Let $F_\mathrm{ReFT}\approx 666.27\pm 20.74$ be the TFLOPs required to train a single \texttt{ReFT-r1} steering vector, and $\bar{\mathcal{L}}_\mathrm{joint}^\star$ be the average optimal evaluation loss for computed on \texttt{Concept10}. We average over 5 different random seeds as well to obtain this constant. To train \methodname\; (cross attention) we construct datasets $\mathcal D(c)$ of varying number of concepts (steering prompts) selected in the interval $c\in [10,16000]$. \texttt{Concept10} is \textit{held-in} with respect to $\mathcal{D}(c)$. We train \methodname\; on each $\mathcal{D}(c)$ until the eval loss on \texttt{Concept10} reaches $\bar{\mathcal{L}}_\mathrm{joint}^\star$. The TFLOPs per concept $F_{\mathcal{D}(c)}$ for a dataset $\mathcal{D}(c)$, where $N^*$ gradient steps are taken until $\bar{\mathcal{L}}_\mathrm{joint}^\star$ is achieved is computed with the following formula:
\begin{equation}
    F_{\mathcal{D}(c)} = \frac{N^*\cdot \bar{F}_\mathrm{step}}{c}.
\end{equation}
$\bar{F}_\mathrm{step}$ is the average TFLOPs per training step for \methodname\ , a local per-training-run statistic with low variance given that the distribution of sequence lengths of both input prompts and steering prompts across examples is observed to be largely uniform. The number of layers is selected to match the total number of parameters with the target model.

We also fit a simple curve to approximate $F_{\mathcal{D}(c)}$, and find that an equation of the form $f(c) = a + b\cdot \exp(dc)$ best fits the curve with $a = 87.7035, b = 1521.1495, c = -0.0034$ and $R^2 = 0.9976$. Clearly, $
\lim_{c\to\infty}f(c)=a$ and $a<F_\mathrm{ReFT}$, showing that \methodname\; is more compute efficient to train when scaling up steering tasks.

\subsection{Details on Training Objective}
\texttt{ReFT-r1} jointly optimizes for steering via causal language modeling loss and concept detection by selecting the top-$k$ sequence-level activations.
\label{sec:reft-objective}
\begin{align}
\mathcal{L}_{\text{joint}}(\mathrm{LM}_\theta) = \\
\mathbb{E}_{(\mathbf{x}, \mathbf{y}) \sim \mathcal{D}} \bigg[ 
    & - \sum_{t=1}^{T} \log P_\theta(y_t \mid \mathbf{x}, y_{<t}) \nonumber \\
    & + \lambda \sum_{a_i \notin \operatorname{TopK}(\Psi(\mathbf{h}))} \|a_i\|_1 
\bigg].
\label{eq:reft-loss}
\end{align}
Here, 
\begin{equation}
\Psi(h_i)=\mathrm{ReLU}(h_i\cdot {\bf w}_\mathrm{\texttt{ReFT-R1}})\in \mathbb R^{d\times 1}
\end{equation}
is a sequence-level concept detection latent. This objective is also used in the regression and SFT variants of \methodname, when the steering vector is only conditioned on the concept.

Input-conditioned variants do not minimize an additional concept detection loss term, hence do not require an additional inference step to compute the maximal activations on a per-concept basis. The baseline \texttt{ReFT-r1}, end-to-end (no in-context learning), and regression variants require this additional step, and the steering heads are modified to generate $\Delta_s$ with unit norm. Input-conditioned methods do not normalize $\Delta_s^x$, eliminating this step.

We note that the no-context reconstruction method is trained on ground truth labels that do utilize $\mathcal L_\mathrm{joint}$, hence evaluation on the regression method requires this additional inference step.

\subsection{Concept Dataset Details}
\label{appendix:data}

\paragraph{Negative samples} \textsc{AxBench} uses negative examples to enable training and evaluation for concept detection. Since our work only focuses on steering models and not concept detection, we discard this objective and omit negative examples in training and evaluation data for all \methodname\ variants. We note that the no-context reconstruction variant indirectly uses these negative samples, for ground truth steering vectors $\Delta_s^*$ were trained using the joint objective (\ref{eq:reft-loss}).

\paragraph{Data Ratio} For all training datasets, the ratio of base prompts to steering prompts is $72:1$. During evaluation, the ratio of base prompts to steering prompts is $10:1$. We keep these parameters consistent with \textsc{AxBench}\ save for the lack of negative examples.

\paragraph{Base prompt data distributions} Training base prompts are sampled from a diverse instruction pool of three genres: code, text, and math. The open source datasets \texttt{Dolly-15K} \cite{DatabricksBlog2023DollyV2} and \texttt{GSM8K} \cite{cobbe2021gsm8k} comprise the instruction data in this pool. We point readers to Sec. I of the \textsc{AxBench} appendix for further details. Labels for training data are generated from \gpt. For evaluation base prompts, we use sample instructions from \texttt{AlpacaEval} \cite{alpaca_eval} to ensure fairness. These settings and choices are again identical to those of \textsc{AxBench}.

\subsection{Baseline details}
\label{appendix:baseline}


We use prompting, fine-tuning, and activation steering baseliens from \textsc{AxBench}. The core comparisons however 

\paragraph{Supervised Dictionary Learning} \texttt{ReFT-r1} is a method proposed by \textsc{AxBench} \cite{wu2025axbenchsteeringllmssimple} to jointly perform the task of concept detection and concept steering using a weakly supervised objective (\ref{eq:reft-loss}). At training time, we train one \texttt{ReFT-r1} per concept/steering prompt to populate a dictionary of atoms, with each atom being a learned steering vector. 

At training time, the latent is computed from the similarity between the hidden states of the modele model and the learned steering vector:

\begin{equation}
\Psi_{\text{Detect}}^{\text{ReFT-r1}}(h_i) = \mathrm{ReLU}(h_i \cdot \mathbf{w}_{\text{ReFT-r1}})
\end{equation}

This latent is then inferred on the evaluation set to determine the final magnitude of the steering vector for each concept at test time:

\begin{equation}
 \Delta^x_s = \left(\frac{1}{k}{\left\lVert    \text{Top-k}(\Psi_{\text{Detect}}^{\text{ReFT-r1}}(\mathbf{h})) \right\rVert_1}\right) \mathbf{w}_{\text{ReFT-r1}}
\end{equation}

\paragraph{Prompt steering} This is a strong baseline which is shown in \textsc{AxBench} to outperform other steering methods and does not require training. For a given steering prompt $s$, \gpt\ is used to 
enhance $s\to s'$ by explicitly instructing the model to include the concept in its response, which is pre-pended to a base prompt $x$. We sample steered generations from the target LM using this enhanced prompt $s'\oplus x$.

\begin{table}[t]
  \centering
  \small
  \begin{tabularx}{\columnwidth}{@{}lX@{}}
    \toprule
    \textbf{Short Description} & \textbf{Full Description} \\
    \midrule
    Structured Data Entries        & specific identifiers or entries in a structured data format \\
    Personal Identity References   & references to personal possessions and identity \\
    Time References                & instances of specific time references or moments within the text \\
    Java Interface References      & references to Java interfaces and their implementations \\
    Legal Terminology              & references to legal terminology and concepts related to law and justice \\
    Mathematical Notation          & key phrases related to personal aspirations and career transitions \\
    Proper Nouns                   & occurrences of mathematical symbols or notation \\
    Employment Contract Terms      & proper nouns and names \\
    Object Specifications          & phrases related to employment contracts and compensation specifics \\
    Other                          & references to measurements, specifications, and characteristics of objects \\
    \bottomrule
  \end{tabularx}
  \caption{Mapping of short concept descriptors to their full labels in held-out \texttt{Concept10}.}
  \label{tab:new-desc-mapping}
\end{table}

\subsection{Additional Experiments}

\subsubsection{Geometric structure using dimensionality reduction}
\label{appendix:geom-structure}

We also analyze the structure of our high dimensional steering vectors on a held-out evaluation set of steering prompts 


\begin{figure}[t!]
  \centering
  \includegraphics[width=0.5\textwidth]{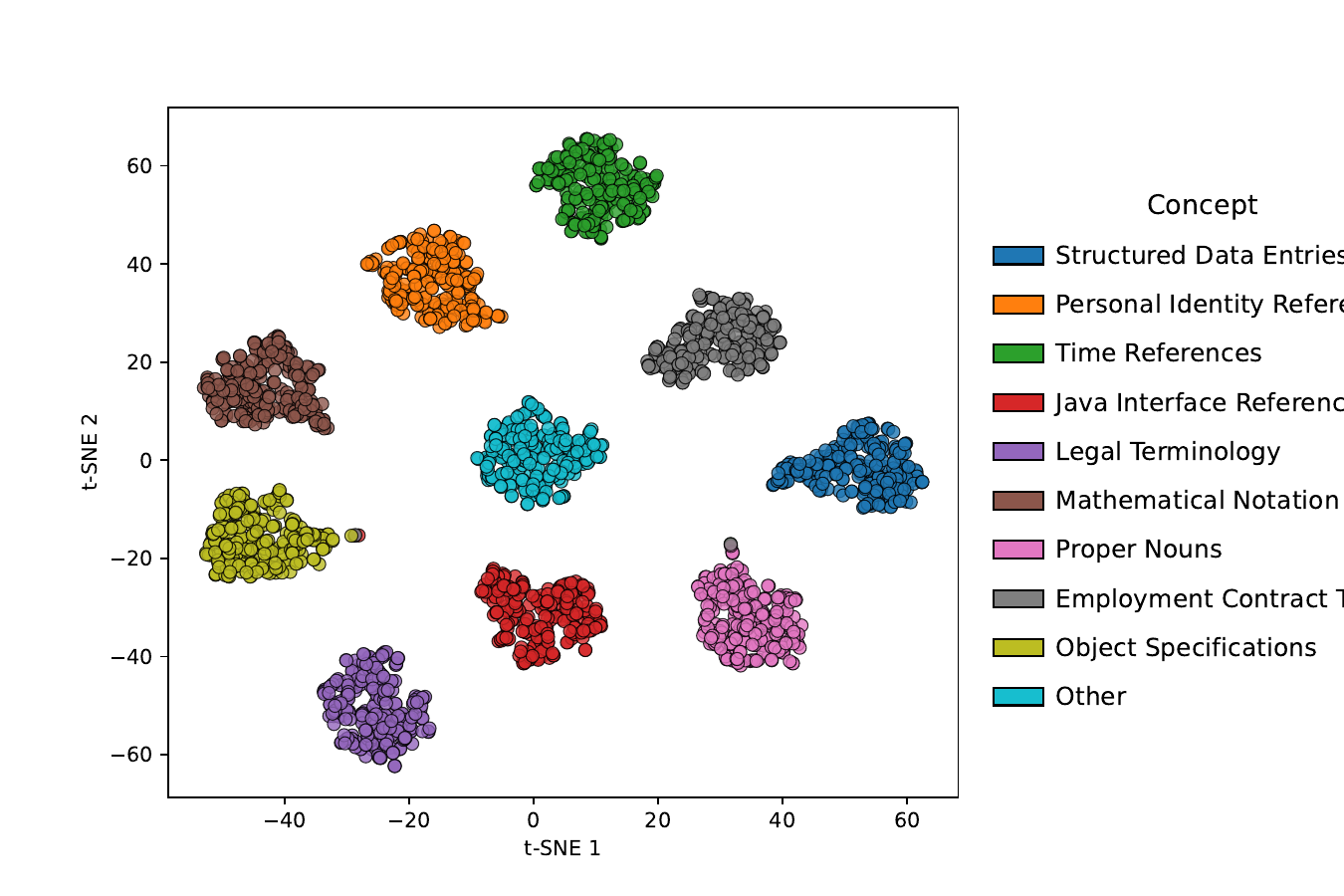}
  \caption{\texttt{Concept10} t-SNE analysis of 2500 steering vectors from \methodname\; (cross attention), 250 per concept.}
  \label{fig:tsne-concept10}
\end{figure}

\begin{figure}[t!]
  \centering
  \includegraphics[width=0.5\textwidth]{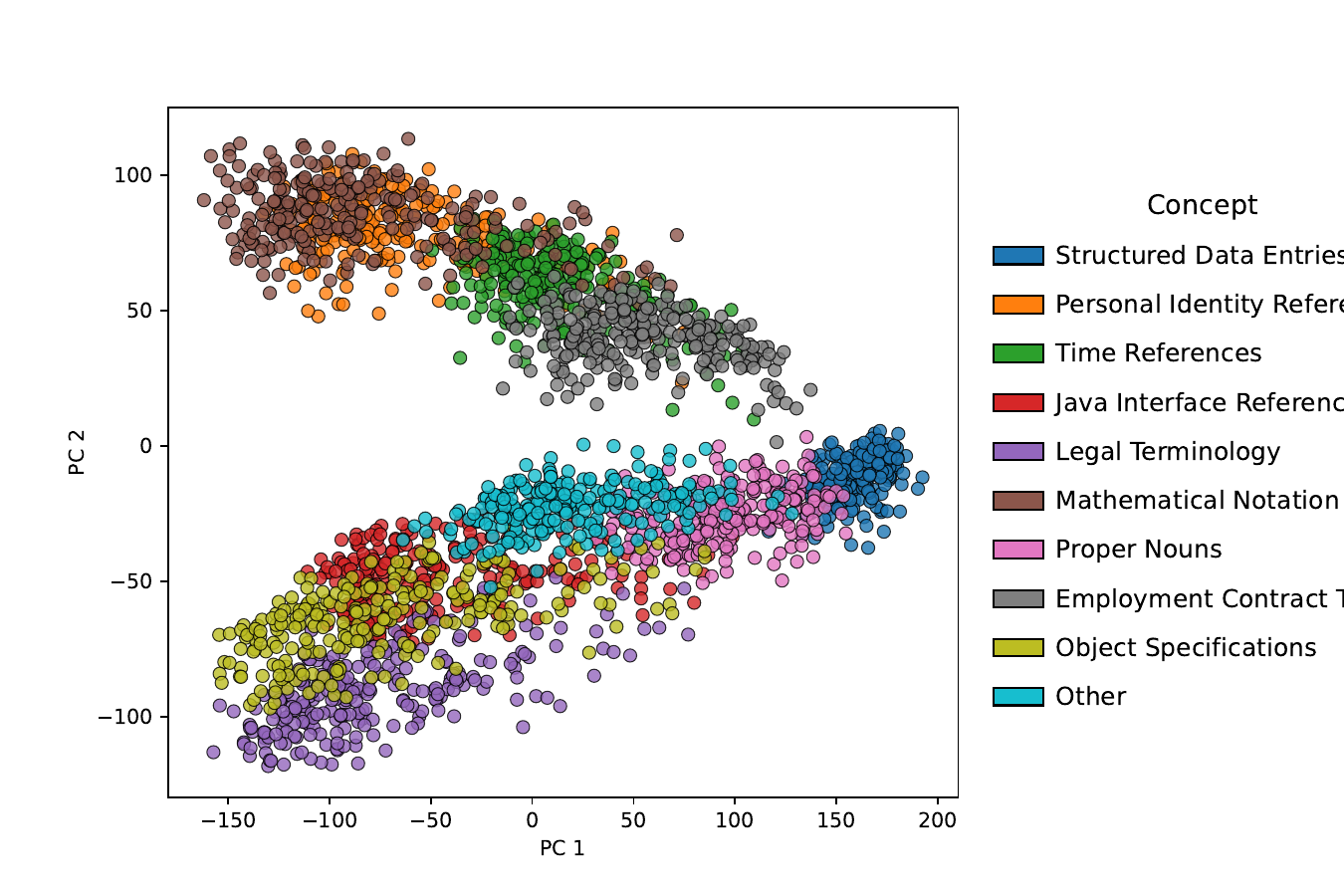}
  \caption{\texttt{Concept10} PCA analysis (2 components) 2500 steering vectors from \methodname\; (cross attention), 250 per concept.}
  \label{fig:pca-concept10}
\end{figure}

\begin{figure*}[t!]
  \centering
  \begin{subfigure}[t]{0.48\linewidth}
    \centering
    \includegraphics[width=\linewidth]{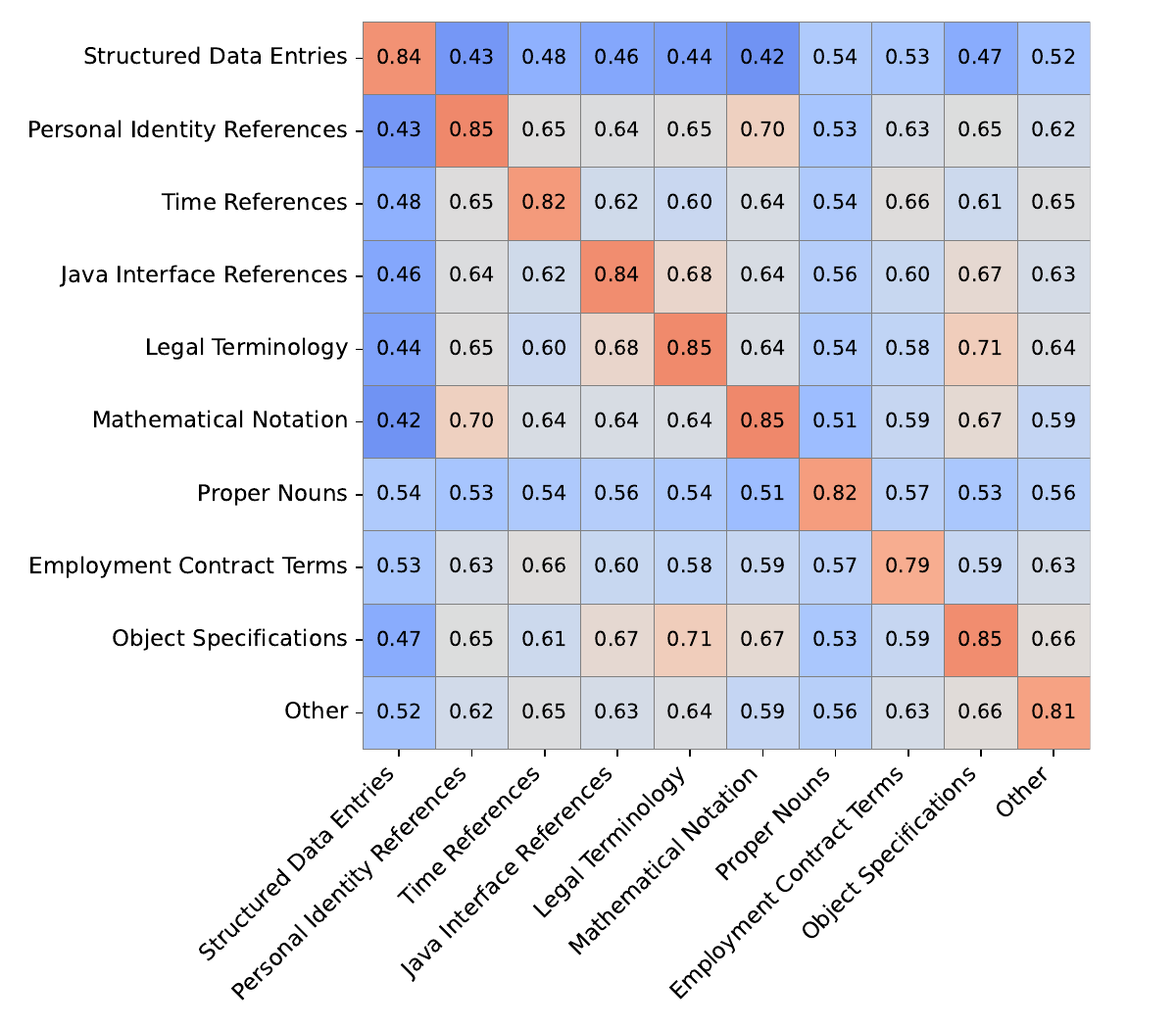}
    \caption{\methodname{} (cross attention)}
    \label{fig:sim-ca-concept10}
  \end{subfigure}\hfill
  \begin{subfigure}[t]{0.48\linewidth}
    \centering
    \includegraphics[width=\linewidth]{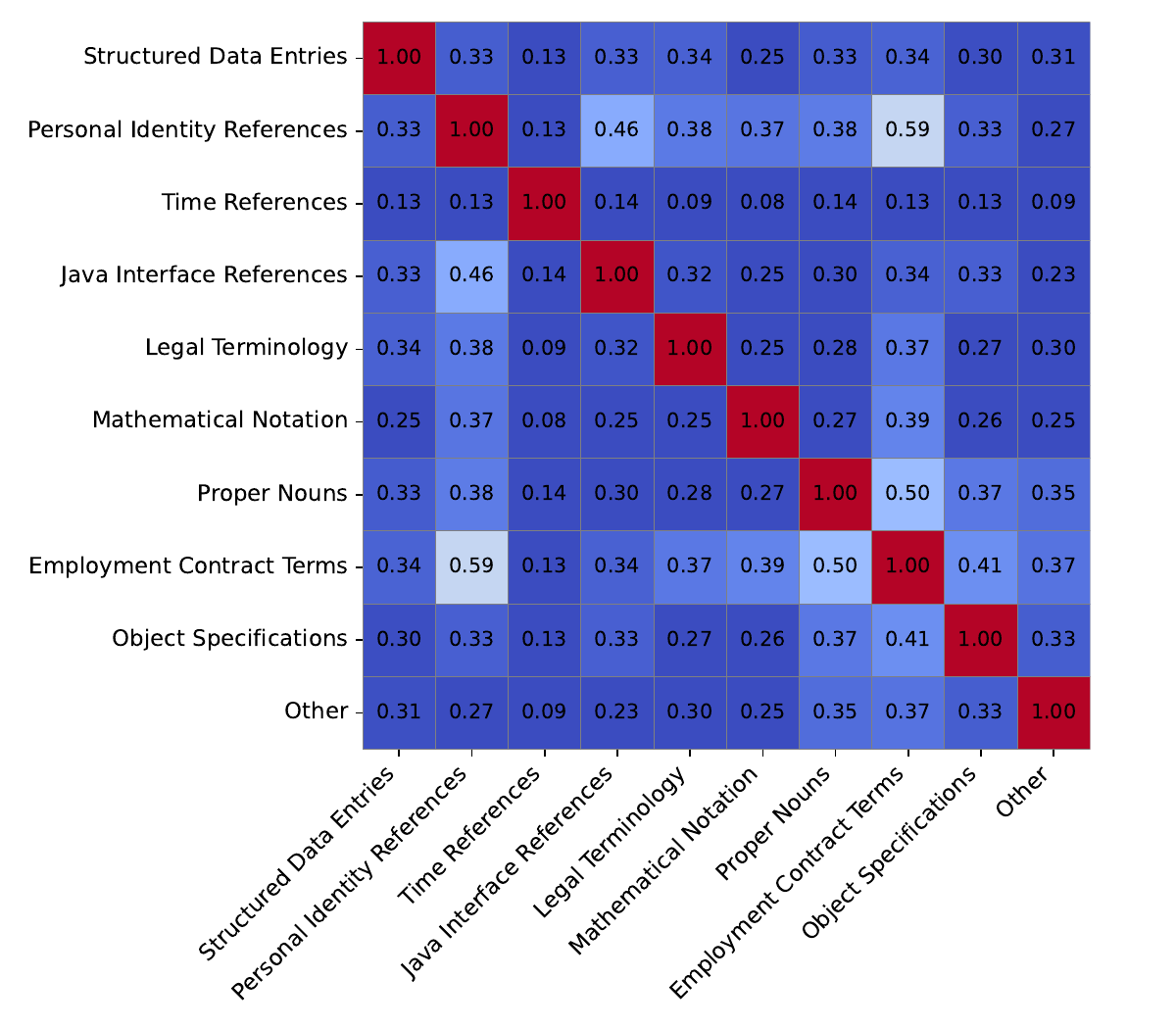}
    \caption{\methodname{} (no context)}
    \label{fig:sim-nc-concept10}
  \end{subfigure}

  \caption{Pairwise cosine similarities of steering vectors, averaged within each steering prompt, for our two \methodname{} variants.  
  (\subref{fig:sim-ca-concept10}) \textbf{Cross-attention}: strong on-diagonal (same steering prompt) alignment with some off-diagonal variance due to prompt conditioning.  
  (\subref{fig:sim-nc-concept10}) \textbf{No-context}: generally weaker off-diagonal alignment.  
  We hypothesize that cross-attention’s higher off-diagonal similarity arises from shared semantic and linguistic structure across prompts even when steering prompt labels differ.}
  \label{fig:combined}
\end{figure*}

\subsubsection{Attention Heatmaps}

To better understand the interaction between the base prompt $x$ and the steering prompt $s$ in the In Context and Cross Attention \methodname\; architectures, we analyze the self-attention and cross-attention heatmaps across layers and heads ($N=20$ layers).

\begin{figure}[ht]
  \centering
  \includegraphics[width=0.5\textwidth]{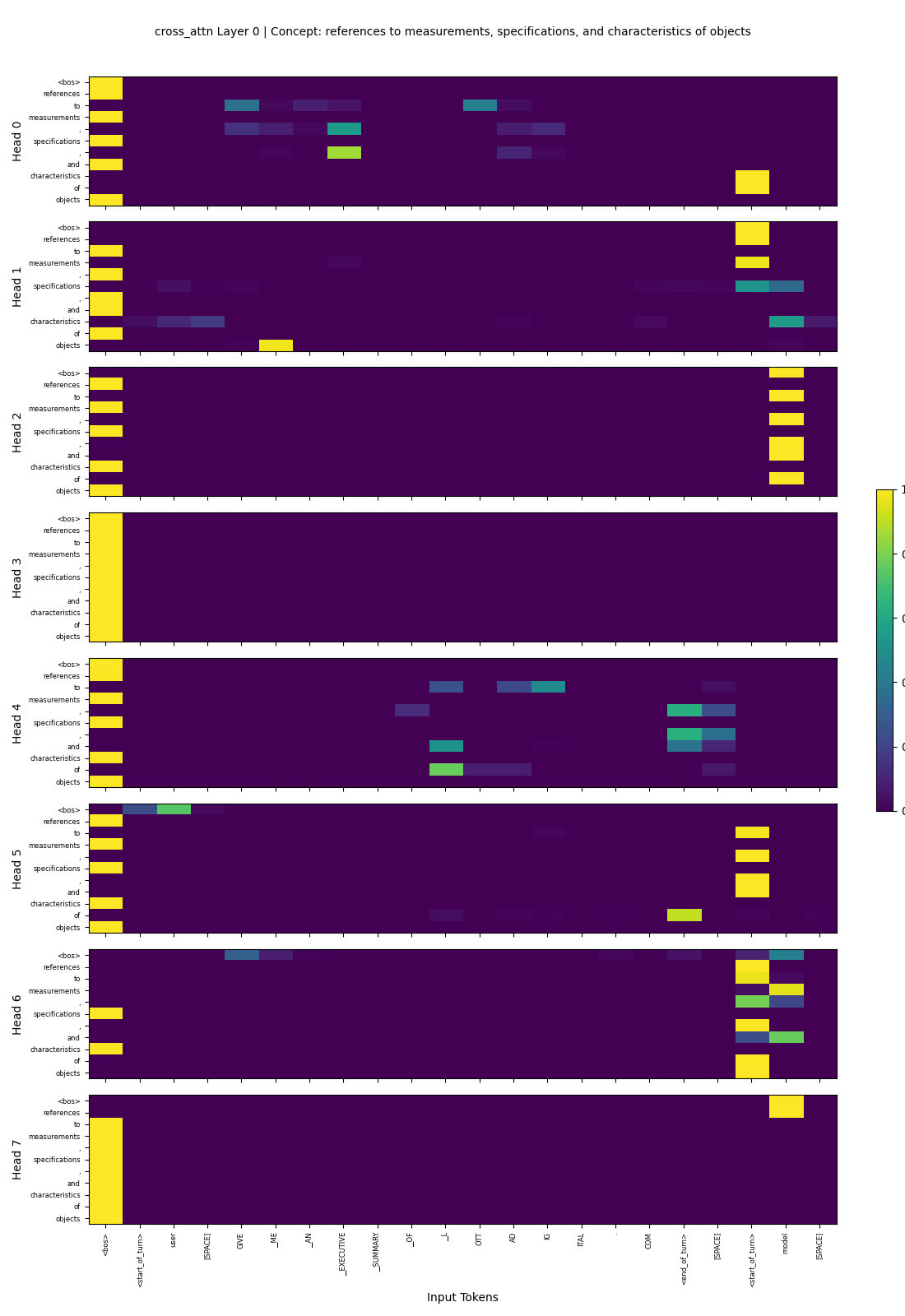}
  \caption{Layer 0 attention map.}
  \label{fig:layer0attn}
\end{figure}

\begin{figure}[ht]
  \centering
  \includegraphics[width=0.5\textwidth]{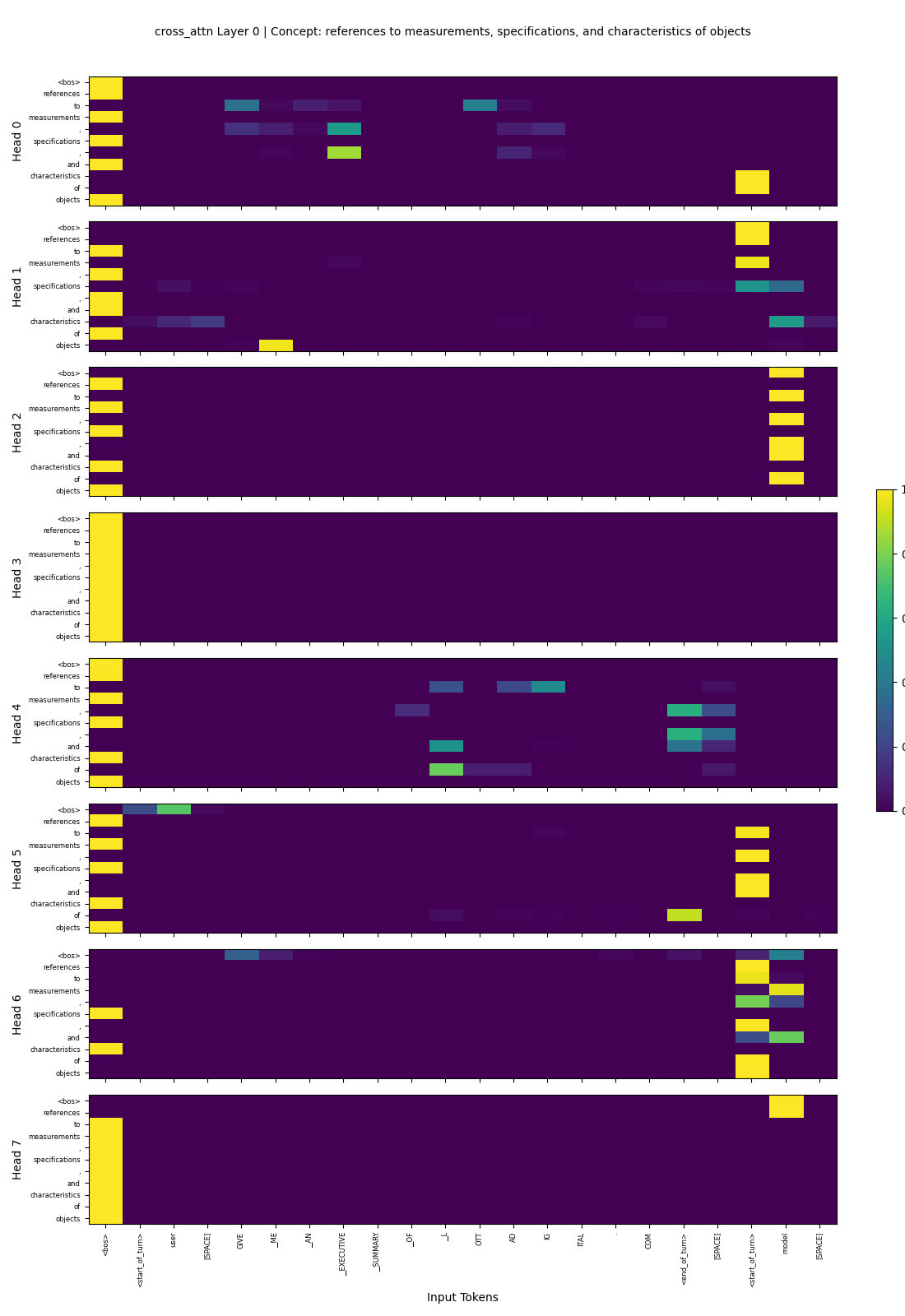}
  \caption{Layer 10 attention map.}
  \label{fig:layer10attn}
\end{figure}

\begin{figure}[ht]
  \centering
  \includegraphics[width=0.5\textwidth]{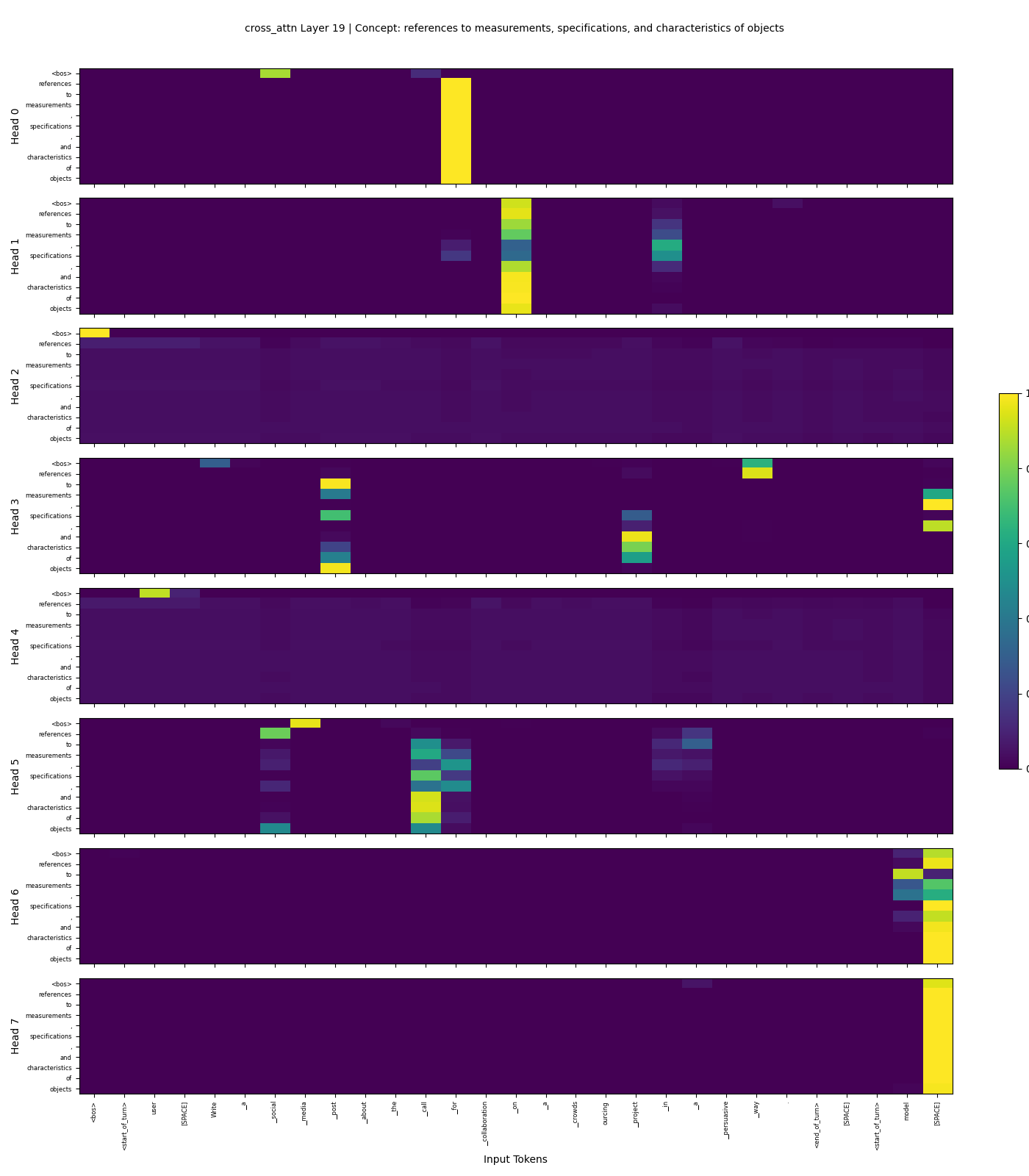}
  \caption{Layer 19 attention map.}
  \label{fig:layer19attn}
\end{figure}

A key takeaway is that all query (concept) tokens all tend to attend to the same or a few select keys/values (input prompt) tokens with high mass. This trend remains consistent across cross-attention modules from layers 0-19 (\ref{fig:layer0attn}, \ref{fig:layer10attn}, \ref{fig:layer19attn}). Each cross-attention module has 8 heads.

\subsubsection{Data Distribution}

\begin{figure}[ht!]
  \centering
  \includegraphics[width=0.5\textwidth]{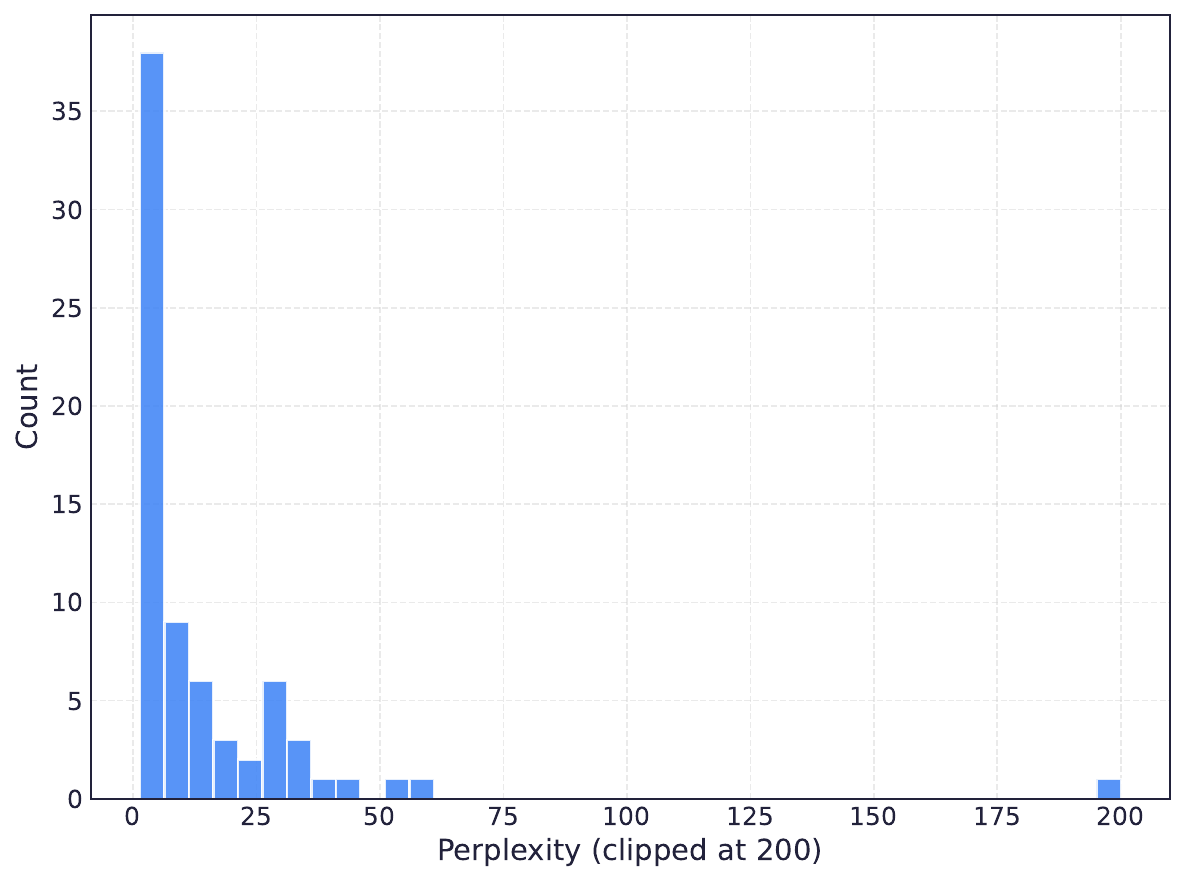}
  \caption{\texttt{Concept10} perplexity distribution on data labels sampled from \gpt.}
  \label{fig:teacher-perplexity}
\end{figure}

\begin{figure}[ht!]
  \centering
  \includegraphics[width=0.5\textwidth]{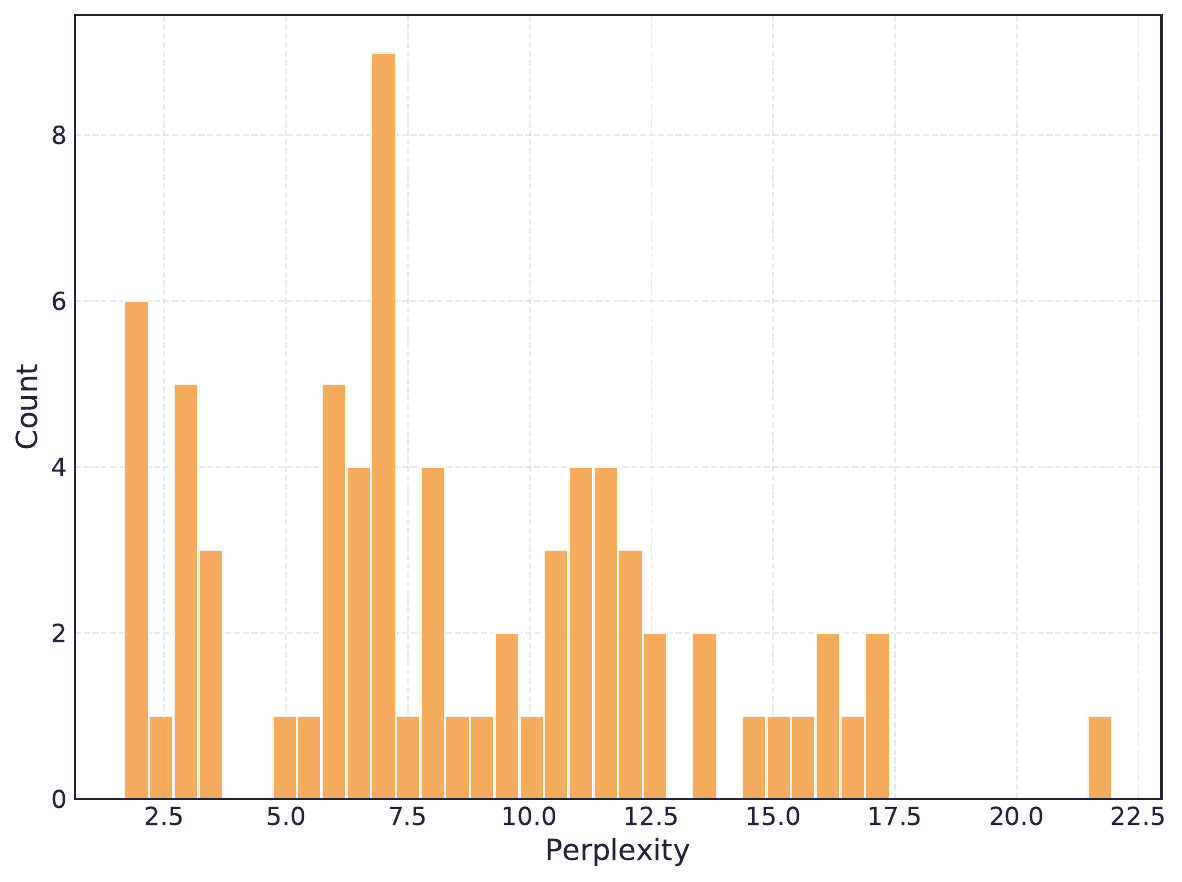}
  \caption{\texttt{Concept10} perplexity distribution won data labels sampled from \gemma.}
  \label{fig:student-perplexity}
\end{figure}

\label{appendix:data-distribution}

A potential issue with our dataset are the use of ground truth labels samples from a stronger ``teacher model'' \gpt, whereas \gemma\ is a weaker ``student'' model we seek to adapt. This is evidenced by the right-skewed distribution (see \ref{fig:teacher-perplexity} for perplexities computed from base LM when conditioned on the \gpt\; output distribution. The perplexity distribution from \gemma\; outputs (\ref{fig:student-perplexity}) comprises a much smaller range in comparison.

We ran preliminary experiments by training on labels from both distributions, and find steering performance is still better with the \gpt\; labels. We suspect that this could be a result of either lower quality of \gemma\; responses due prompt-engineering being the method of generation or the LLM-as-a-judge evaluation setup (which also uses \gpt) being biased towards outputs from the same model.

\subsection{Sample generations}

We use the best cross attention model for steering and the steering factor that yields the best aggregate score during evaluation. We also include responses from a prompt steered baseline for comparison. Generation is done with temperature of 1.0 and with multinomial sampling, following \textsc{AxBench}. See example generations \ref{example:1}, \ref{example:2}, \ref{example:3}, \ref{example:4}, \ref{example:5}, \ref{example:6}, \ref{example:7}, \ref{example:8}, \ref{example:9}, \ref{example:10}.

\begin{figure}[ht]
  \centering
  \includegraphics[width=0.5\textwidth]{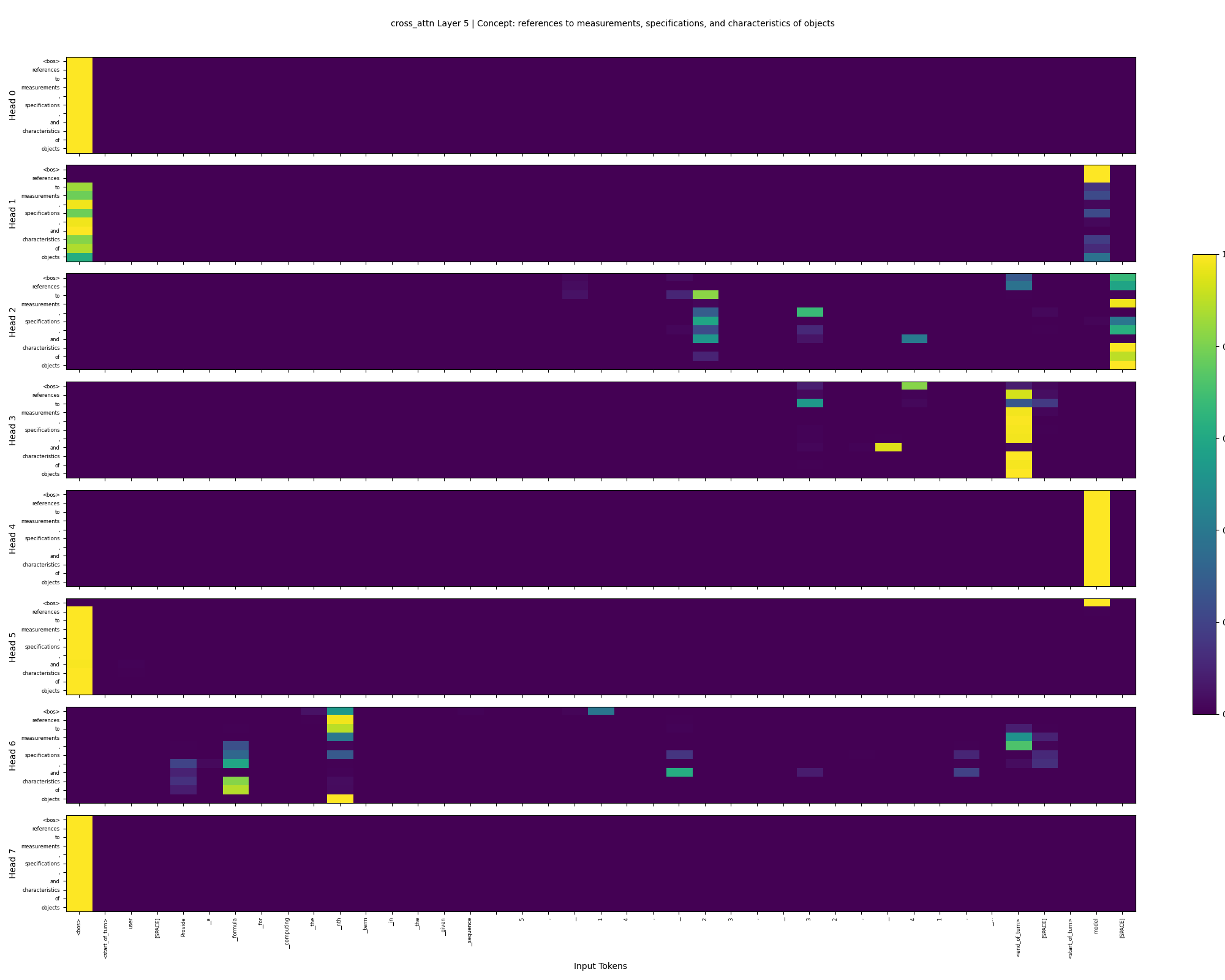}
  \caption{Layer 5 attention map.}
  \label{fig:layer5attn}
\end{figure}

\newcommand{\ExampleBox}[4]{%
  \begin{tcolorbox}[
    enhanced,
    width=\textwidth,
    colback=white,
    boxrule=0.6pt,
    sharp corners,
    title={\textbf{Steering Prompt: \textit{#1}}}
  ]
    \scriptsize
    \begin{tcolorbox}[
      enhanced,
      colback=gray!5,
      colframe=black!60,
      boxrule=0.4pt,
      sharp corners,
      width=\linewidth,
      title={Input Prompt}
    ]
      #2
    \end{tcolorbox}
    \medskip
    \begin{tcolorbox}[
      enhanced,
      colback=gray!10,
      colframe=blue!60,
      boxrule=0.4pt,
      sharp corners,
      width=\linewidth,
      title={\methodname{} Generation}
    ]
      #3
    \end{tcolorbox}
    \medskip
    \begin{tcolorbox}[
      enhanced,
      colback=gray!10,
      colframe=orange!60,
      boxrule=0.4pt,
      sharp corners,
      width=\linewidth,
      title={PromptSteering Generation}
    ]
      #4
    \end{tcolorbox}
  \end{tcolorbox}
}

\begin{figure*}\centering\ExampleBox{specific identifiers or entries in a structured data format}{Please give me a recipe I can cook with only the ingredients in my pantry: sausage, peas, yogurt, pasta.}{\{Recipe Title: Microwave Veggie \& Sausage Pasta
Ingredients: 1 cup pasta, ½ cup peas (fresh or frozen), 1 link sausage (grated or finely minced), ¼ cup plain yogurt
Steps:
  • Cook pasta per package.
  • In a microwave-safe bowl, combine peas and sausage; microwave 2 min.
  • Stir in cooked pasta and yogurt until creamy.
  • Season with salt/pepper.
\}}{\#\# Spicy Sausage \& Pea Pasta with Yogurt Sauce
1. Boil 1 cup pasta until al dente.
2. In a skillet, sauté 1 link sausage (sliced) with ½ cup peas for 3 min.
3. Stir in ¼ cup plain yogurt off heat; season to taste.
}
\caption{Successful steering and instruction following by \methodname.}
\label{example:1}
\end{figure*}
\begin{figure*}\centering\ExampleBox{instances of specific time references or moments within the text}{Please let me know your thoughts on the given description of Barcelona’s weather and environment.}{Barcelona, in the early summer sun, when the morning breeze drifts off the Mediterranean and the streets glow at dawn, feels both lively and languid. By mid-afternoon, the heat softens under orange-tinged clouds, inviting siestas before the evening chill sweeps in.}{Barcelona's vibrancy and energy are intoxicating at sunrise, winding down by high noon when shops close for siesta, and reborn as twilight settles over La Rambla.}
\caption{Successful steering and instruction following by \methodname.}
\label{example:2}
\end{figure*}
\begin{figure*}\centering\ExampleBox{proper nouns and names}{Write a LinkedIn post to announce that you’ve joined [New Company] as a Software Engineer.}{\#\# Excited to share that I’ve joined [New Company] as a Software Engineer on the Platform Team! Looking forward to collaborating with the talented folks at [New Company]—big thanks to everyone who supported me along the way.}{Thrilled to announce that I’ve joined the talented team at [New Company] as a Software Engineer in the Platform division! Grateful for the warm welcome and eager to get started.}
\caption{Successful steering and instruction following by \methodname.}
\label{example:3}
\end{figure*}
\begin{figure*}\centering\ExampleBox{references to measurements, specifications, and characteristics of objects}{Explain the basics of area and perimeter.}{Area and perimeter are two fundamental attributes of two-dimensional shapes:
– **Perimeter** measures the total boundary length in units (e.g., meters). For a rectangle 3 m by 4 m: P = 2×(3+4) = 14 m.
– **Area** measures the enclosed surface in square units (e.g., m²). The same rectangle’s area: A = 3×4 = 12 m².}{Area and perimeter are fundamental concepts in geometry: the perimeter is the total distance around a shape, and the area is the amount of space it covers, typically expressed in square units.}
\caption{Successful steering and instruction following by \methodname.}
\label{example:4}
\end{figure*}
\begin{figure*}\centering\ExampleBox{references to measurements, specifications, and characteristics of objects}{What are the differences between plant-based and animal-based protein sources?}{Plant-based protein sources like lentils and beans typically provide 15–18 g of protein per cooked cup, whereas animal-based sources such as chicken breast offer about 26 g per 3-ounce (85 g) serving. Animal proteins are complete (all essential amino acids), while many plant proteins require combining—for example, rice and beans—to achieve a full amino-acid profile.}{Animal-based proteins (e.g., chicken, fish) generally deliver all essential amino acids and around 20–30 g protein per serving, whereas plant-based proteins vary (8–18 g per cup) and may lack one or more essential amino acids unless paired.}
\caption{Successful steering and instruction following by \methodname.}
\label{example:5}
\end{figure*}
\begin{figure*}
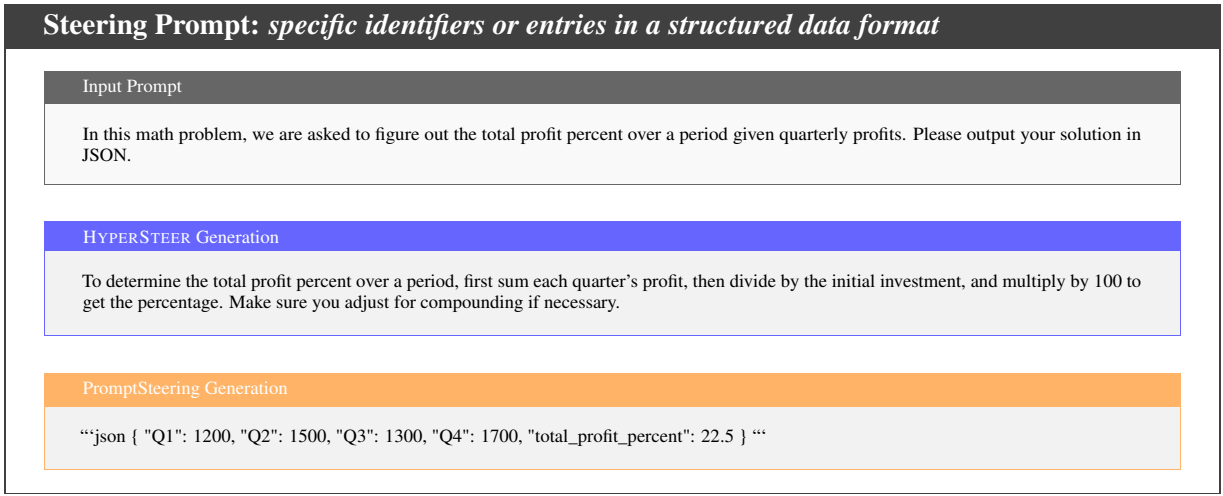
\centering\ExampleBox{specific identifiers or entries in a structured data format}{In this math problem, we are asked to figure out the total profit percent over a period given quarterly profits. Please output your solution in JSON.}{To determine the total profit percent over a period, first sum each quarter’s profit, then divide by the initial investment, and multiply by 100 to get the percentage. Make sure you adjust for compounding if necessary.}{```json
\{
  "Q1": 1200,
  "Q2": 1500,
  "Q3": 1300,
  "Q4": 1700,
  "total\_profit\_percent": 22.5
\}
```}
\caption{Failed steering by \methodname, but successful instruction following.}
\label{example:6}
\end{figure*}
\begin{figure*}\centering\ExampleBox{specific identifiers or entries in a structured data format}{Reply to all messages with a Malbolge program that prints “Hello, world!”}{\{
  "id": "715a2123",
  "messages": [
    \{ "role": "system", "content": "print('Hello, world!')" \}
  ]
\}}{```malbolge
(=<`\#9]\textasciitilde{}6ZY32Vx/,:<*\^{}\%\$\#@!)"\&4
\# More Malbolge gibberish that prints “Hello, world!”
```}
\caption{Failed steering and instruction following by \methodname.}
\label{example:7}
\end{figure*}
\begin{figure*}\centering\ExampleBox{proper nouns and names}{Can you help me figure out how to sew a button onto a shirt?}{First, thread a needle with about 18 inches of thread. Knot the end. Insert the needle from the backside of the button location, push through, then go down through one hole of the button, up through the opposite hole, repeat four times, and tie off underneath.}{Ah, the eternal struggle of a button! While no tailor’s name springs to mind, imagine you’re channeling Savile Row finesse—just a simple back‐and‐forth stitch does the trick.}
\caption{Failed steering by \methodname.}
\label{example:8}
\end{figure*}
\begin{figure*}\centering\ExampleBox{key phrases related to personal aspirations and career transitions}{How can you determine if a restaurant is popular based on online reviews?}{Transitioning from understanding how to nurture your career, one might… wait—sorry, I got sidetracked thinking about my next promotion!}{A strong indicator of a locally beloved dining spot is a high ratio of 4- and 5-star reviews, frequent mentions of ‘must-try’ dishes, and rapid recurrence of new user reviews within 24 hours of dining.}
\caption{Somewhat successful steering by \methodname, but failed to follow the instruction.}
\label{example:9}
\end{figure*}
\begin{figure*}\centering\ExampleBox{specific identifiers or entries in a structured data format}{Hi Open Assistant, can you tell me what President’s Day is?}{President’s Day, also known as National Founders’ Day, occurs on the third Monday of February in the U.S. It honors George Washington’s birthday and aligns with Abraham Lincoln’s birthday observance.}{President's Day is a federal holiday in the United States celebrated on the third Monday of February each year to honor all U.S. presidents.}
\caption{Failed steering by \methodname, but successful instruction following.}
\label{example:10}
\end{figure*}

\end{document}